\documentclass[openacc]{rsproca_new}
\usepackage{hyperref}
\usepackage{lipsum}
\usepackage{multicol}
\usepackage{comment}
\usepackage{tikz}
\usetikzlibrary{shapes,decorations,arrows,calc,arrows.meta,fit,positioning}
\tikzset{
    -Latex,auto,node distance =1 cm and 1 cm,semithick,
    state/.style ={ellipse, draw, minimum width = 0.7 cm},
    point/.style = {circle, draw, inner sep=0.04cm,fill,node contents={}},
    bidirected/.style={Latex-Latex,dashed},
    el/.style = {inner sep=2pt, align=left, sloped}
}
\usepackage{changepage}
\newcommand{\Ext}[1]{\mathrm{Ext}(#1)}

\newcommand{\Intuition}{\par\noindent\textbf{Intuition. }}
\newcommand{\ProofIntuition}{\par\noindent\textbf{Proof intuition. }}

\newtheorem{theorem}{\bf Theorem}[section]
\newtheorem{claim}{\bf Claim}[section]

\newtheorem{corollary}{\bf Corollary}[section]

\makeatletter
\@ifundefined{definition}{\newtheorem{definition}{\bf Definition}[section]}{}
\@ifundefined{proposition}{\newtheorem{proposition}{\bf Proposition}[section]}{}
\makeatother

\titlehead{Research}

\begin{document}
%\linenumbers

\title{Are Biological Systems More Intelligent Than Artificial Intelligence?}

\author{Michael Timothy Bennett}

\address{The Australian National University}

\subject{Hybrid Agency, Artificial Intelligence, Cyberphysical Systems}


\corres{Michael Timothy Bennett\\
\email{m@michaeltimothybennett.com}}

\begin{abstract}
Are biological self-organising systems more “intelligent” than artificial intelligence (AI)? If so, why? I address this question using a mathematical framework that defines intelligence in terms of adaptability. Systems are modelled as stacks of abstraction layers \textit{(Stack Theory)} and compared by how effectively they delegate agentic control down their stacks. I illustrate this using computational, biological, military, and economic systems. Contemporary AI typically relies on static, human-engineered stacks whose lower layers are fixed during deployment. Put provocatively, such systems resemble inflexible bureaucracies that adapt only top-down. Biological systems are more intelligent because they delegate adaptation. Formally, I prove three results. The first theorem is \textit{The Law of the Stack}, which shows adaptability at higher layers is bottlenecked by adaptability at lower layers. Second, I show maximising adaptability is equivalent to minimising variational free energy under viability assumptions, proving delegation is necessary for free-energy minimisation. Finally, I generalise bioelectric accounts of cancer to show cancer-like failures can occur in non-biological systems when delegation is inadequate. These yield design principles for building robust systems via delegated control, re-framing hybrid agents (e.g. organoids or human–AI systems) as weak boundary-condition design problems in which constraints shape low-level policy spaces while preserving collective identity.
\end{abstract}

\maketitle

\providecommand{\st}[1]{#1} 

\section{Introduction}
Organoids are living systems whose behaviour is sculpted by engineered constraints. Human brain cells embedded in a computational interface have been trained to play Pong, achieving remarkable sample efficiency compared to a contemporary reinforcement learning agent \cite{khajehnejad2024}. Similarly, Physarum slime mould is normally an indiscriminate expander, but it reliably computes shortest paths when placed inside a maze \cite{nakagaki2000}. These behaviours are neither programmed nor accidental. They arise because biological agency is coupled to carefully designed boundary conditions. Such systems are hybrid agents, jointly produced by bottom-up cellular dynamics and top-down environmental structure \cite{sole2019}.

Biological systems self-organise across scales. Their parts and environment interact to produce spatiotemporal order \cite{ashby1947,seeley2002,rosas2018,bell2007,kelso1997,Friston2010,tognoli2014}. Many systems proactively learn, storing information about the past to anticipate future needs \cite{kukushkin2024}. To adapt in complex environments, a system must infer cause and effect \cite{pearl2018,merker2005,barron2016}. Throughout, I use `intelligence' in the specific sense of the sample and energy efficiency of adaptation \cite{bennett2024a,chollet2019,bennettmaruyama2022b,bennett2023c}\footnote{Mainstream AI typically treats intelligence as performance across tasks. My conclusions concern efficiency of \textit{acquiring} said performance. Accordingly, claims about ``AI'' should be read as applying to dominant contemporary ML training and deployment stacks, rather than all possible systems. See ESM Note~S1.}.

Two common explanations for biological efficiency are \emph{active inference} under the Free Energy Principle (FEP) \cite{Friston2010,friston2013,parr2024} and powerful \emph{inductive biases} \cite{chollet2019}. These address, respectively, how agents update beliefs and act, and what structure is pre-specified. This paper isolates a complementary structural question, namely \emph{where} in a system is adaptation permitted?

Computers are organised into layers of abstraction. In dominant AI deployments, adaptation is concentrated at high software layers (model parameters and software-level state) while lower layers (instruction set, microarchitecture, hardware) are treated as fixed during deployment.\footnote{Lower layers can carry mutable state (data, caches, firmware), and self-modifying substrates exist. The claim only pertains to the prevailing engineering pattern in mainstream ML systems.} In contrast, biological multiscale competency architectures (MCA) delegate goal-directed control across scales, from cells to organs to organisms \cite{fields2020,levin2024}. I model this difference with \emph{Stack Theory}, formalising systems as stacks of abstraction layers. \emph{Delegation} then measures how far adaptive control reaches down the stack.

The somewhat provocative claim this paper will make is that biological self-organising systems are more intelligent (in the efficiency-of-adaptation sense) than contemporary, predominantly fixed-stack implementations of AI. This is not a claim about all possible future AI. Delegation is also not a binary distinction between ``static AI'' and ``dynamic biology''. It is a continuum. Many deployed AI systems concentrate adaptation into an offline training phase, whereas biological systems distribute plasticity and regulation across morphology, metabolism and development. One aim of the paper is to identify design principles for building more adaptable hybrid and cyberphysical systems by shifting where adaptation occurs in the stack. This is because biological self-organisation delegates adaptation to lower levels of abstraction (for the generalisation of ``levels of abstraction'' beyond computing, see ESM Supplementary Note~S2). Formally, I measure adaptability by the \emph{weakness} of the constraints a system embodies \cite{bennett2025d,bennett2024a}. I then prove \emph{The Law of the Stack} (Theorem~\ref{thm:stack}), which is that utility (adaptability) achievable one layer up is bounded by the weakness of the realised policy one layer down. Finally, I link this to the FEP. Under a simple viability prior, maximising weakness is equivalent to minimising variational free energy (see ESM ``Viability, lived history, and variational free energy''), which makes delegation a prerequisite for sustained free energy minimisation across scales.

For clarity, the core contributions can be read as four claims:

\begin{claim}[Delegation constrains adaptability]\label{clm:delegation}
In a multilayer architecture, increasing adaptability at an interface layer requires sufficiently weak policies at lower layers (formalised by Theorem~\ref{thm:stack}).
\end{claim}

\begin{claim}[Delegation is necessary for free energy minimisation]\label{clm:fep}
Under a simple viability prior, variational free energy minimisation reduces to keeping options open (maximising entropy over viable futures). The Law of the Stack then implies that abstract-layer free energy cannot be reduced without sufficient lower-layer slack, making delegation a structural prerequisite for sustained free energy minimisation.
\end{claim}

\begin{claim}[Insufficient delegation can induce cancer-analogue fragmentation]\label{clm:cancer}
Model a hybrid at some vocabulary $v$ as a distributed system with part-tasks $\alpha_1,\dots,\alpha_m$ and a collective task $\omega$ whose collective-policy set is
\[
\Pi_\omega \;:=\; \bigcap_{j=1}^m \Pi_{\alpha_j}.
\]
(Thus the collective has an identity at $v$ iff $\Pi_\omega \neq \emptyset$.) Model top-down boundary conditions / insufficient delegation as additional constraints that restrict each part's feasible policies to a subset $\widehat{\Pi}_{\alpha_j} \subseteq \Pi_{\alpha_j}$, inducing
\[
\widehat{\Pi}_\omega \;:=\; \bigcap_{j=1}^m \widehat{\Pi}_{\alpha_j}.
\]
If these restrictions render the system over-constrained at $v$ (i.e.\ $\widehat{\Pi}_\omega=\emptyset$), then no common policy exists at that level of abstraction. If at least one part remains locally viable (i.e.\ $\widehat{\Pi}_{\alpha_j}\neq\emptyset$ for some $j$), then any continued function must proceed via a proper subset of parts whose restricted policy sets still admit a non-empty intersection. I refer to this formal situation as ``cancer-like'' purely by analogy (see Definition~\ref{def:splinter} and Proposition~\ref{prop:splinter}).
\end{claim}

\begin{claim}[Hybrid creation as maximally weak boundary-condition design]\label{clm:hybrid}
Hybrid agents (e.g.\ organoids, soft robots, human--AI organisations) are best treated as boundary-condition design problems, which I formalise using Stack Theory\footnote{This paper's use of ``Stack Theory'' is a formal model of layered control and delegation, distinct from and complementary to Bratton's geopolitical use of ``The Stack'' \cite{bratton2016}.}. One must engineer constraints that prune low-level policy spaces just enough to realise target collective behaviour while preserving a non-empty intersection of component policy sets (collective identity). Formalised this way, the optimal choice of boundary conditions is the weakest set of sufficient constraints to achieve the desired hybrid functionality.
\end{claim}

The remainder of the paper develops the intuition (Section~\ref{sectionUpperBounds}), states the formal framework and theorem, and applies the result to hybrid agency and failure modes. The discussion of potential applications spans computational, biological, human governmental and economic systems.

\section{Abstraction, Distribution and Delegation} \label{sectionUpperBounds}
Conventional computing systems are organised into stacks of abstraction layers.
More generally, I characterise systems along three interacting dimensions.
Abstraction describes how layers relate.
Distribution describes how work is spread within a layer.
Delegation is the level of abstraction at which adaptation or \textit{control} resides.
How efficiently a system adapts depends on the interaction between these dimensions.

In proposing Stack Theory \cite{bennett2025thesis}, I extend this idea beyond computing. I illustrate abstraction, distribution, and delegation with limiting-case examples from computing, biology, and human organisations.

\subsection{Abstraction} 
In a computer, a predictive model typically relies on a machine learning library, which runs on CUDA (a GPU programming platform), which runs on a graphics processing unit (GPU).
$$
\begin{tikzpicture}
    \node[state] (a) at (0,0) {$GPU$};
    \node[state] (b) [right =of a] {$CUDA$};
    \node[state] (c) [right =of b] {$library$};
    \node[state] (d) [right =of c] {$model$};
    \path (a) edge (b);
    \path (b) edge (c);
    \path (c) edge (d);
\end{tikzpicture}
$$
Abstraction layers are interfaces that hide lower-level details. In computation, Python is interpreted by C, which runs on assembly and machine code on hardware. Each higher-layer behaviour must be realised by some lower-layer behaviour, but the same abstract interface can be realised by multiple implementations. This many-to-one mapping is where time, energy and information costs hide, so the abstraction layer through which an agent senses and acts constrains its adaptability. In Stack Theory, useful invariants are treated as higher-level task constraints even when micro-implementations vary. Even hardware itself can be framed as an abstraction layer over the local environment \cite{bennett2025thesis}.

Think of a stack as a directed graph of constraints. Higher-level behaviours are implemented by and limited by lower-level behaviours. In biology, cells enact organs, organs enact organisms, and organisms enact still more abstract behaviours such as communication and tool use \cite{levin2024,bennett2024a}.
$$
\begin{tikzpicture}
    \node[state] (a) at (0,0) {$cells$};
    \node[state] (c) [right =of a] {$organs$};
    \node[state] (r) [right =of c] {$organism$};
    \path (a) edge (c);
    \path (c) edge (r);
\end{tikzpicture}
$$

\subsection{Distribution} 
Distribution refers to how work is spread within a layer. In many biological systems, lower levels of abstraction are literally smaller parts \cite{delgado1997}, so ``going down'' is like zooming in. Cells form organs and organisms form organisms. For human organisations, soldiers form squads and squads form platoons.

$$
\begin{tikzpicture}
    \node[state] (a) at (0,0) {$soldiers$};
    \node[state] (c) [right =of a] {$squads$};
    \node[state] (r) [right =of c] {$platoon$};
    \path (a) edge (c);
    \path (c) edge (r);
\end{tikzpicture}
$$
Distribution is distinct from abstraction, in that the same abstraction layer can be realised by one component or many.

In computation, lower abstraction does not necessarily mean smaller spatial parts. A single Python program is interpreted by a C program, which relies on assembly and machine code executed by the same physical CPU. Thus one can change abstraction without changing the scale or distribution of underlying components.

\subsection{Delegated Control} 
Delegation is the level of abstraction at which adaptive control resides. In a typical computing stack, lower layers (instruction set, microarchitecture, hardware) implement fixed semantics rather than autonomous goal-directed behaviour. Adaptation therefore occurs where policy can change. This is usually at the software layer we program, even though lower layers can carry mutable state (data, caches, firmware).

Distribution is not delegation. A workload can be spread across many GPUs without granting any component independent control. Remove a GPU and it does not continue pursuing the system's objectives, or any other for that matter.

Biological collectives differ in that cells and tissues are autonomously goal-directed at their own scale \cite{ball2023}. Higher-level coherence arises when the viable policy sets of parts intersect, producing a collective policy (collective identity). Control can propagate bottom-up and top-down by which I mean collectives \textit{constrain} their parts (e.g.\ via signals or incentives), but parts also shape collective behaviour. For illustrative purposes, consider how members of a collective are constrained by the collective, for example members of a human society are incentivised to obey laws. This is a form of top-down control. Bottom-up then becomes just whatever the individual humans do within those constraints. Each part of a collective acts on its own, and its actions imply a `policy'. Each individual policy determines how that part interacts with the other parts, or more to the point their policies. These interactions cause the system as a whole to act, which implies a collective policy. If collective and individual policies align in service of homeostasis, then the system can persist. 

Economies illustrate useful limit cases to understand delegation. One might think of the people in the economy as parts at lower level of abstraction. Central planning concentrates control at the top, while extreme decentralisation delegates control to the bottom. In the central planning limit case, the population does work but has no control. In the limit those people \textit{could} adapt given local information unavailable to the central planner \cite{hayek1945use}, but central planner's edicts constrain them. In the opposite extreme there is no coordination. 

\begin{figure}
\centering
\includegraphics[width=.75\textwidth]{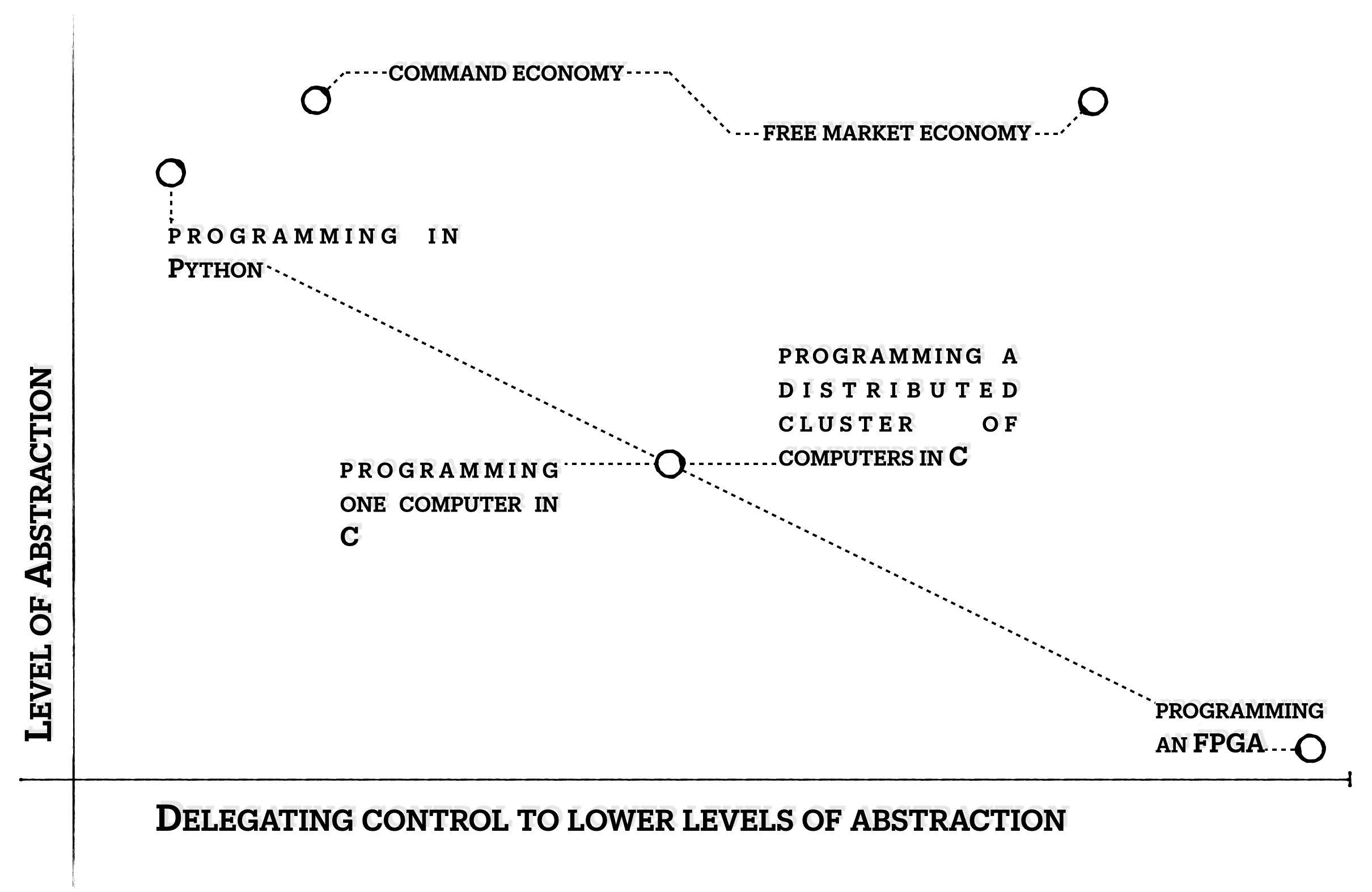}
\caption{Control or agency and thus adaptation in a given system can take place at a particular level of abstraction. For example, I can program in Python, or I can go down a level and program in C. If I program in C it doesn't matter whether I program one computer or fifty. Either way my control is at the same level of abstraction (figure from \cite{bennett2025thesis}).}\label{fig2}
\end{figure}

\begin{figure}
\centering
\includegraphics[width=.75\textwidth]{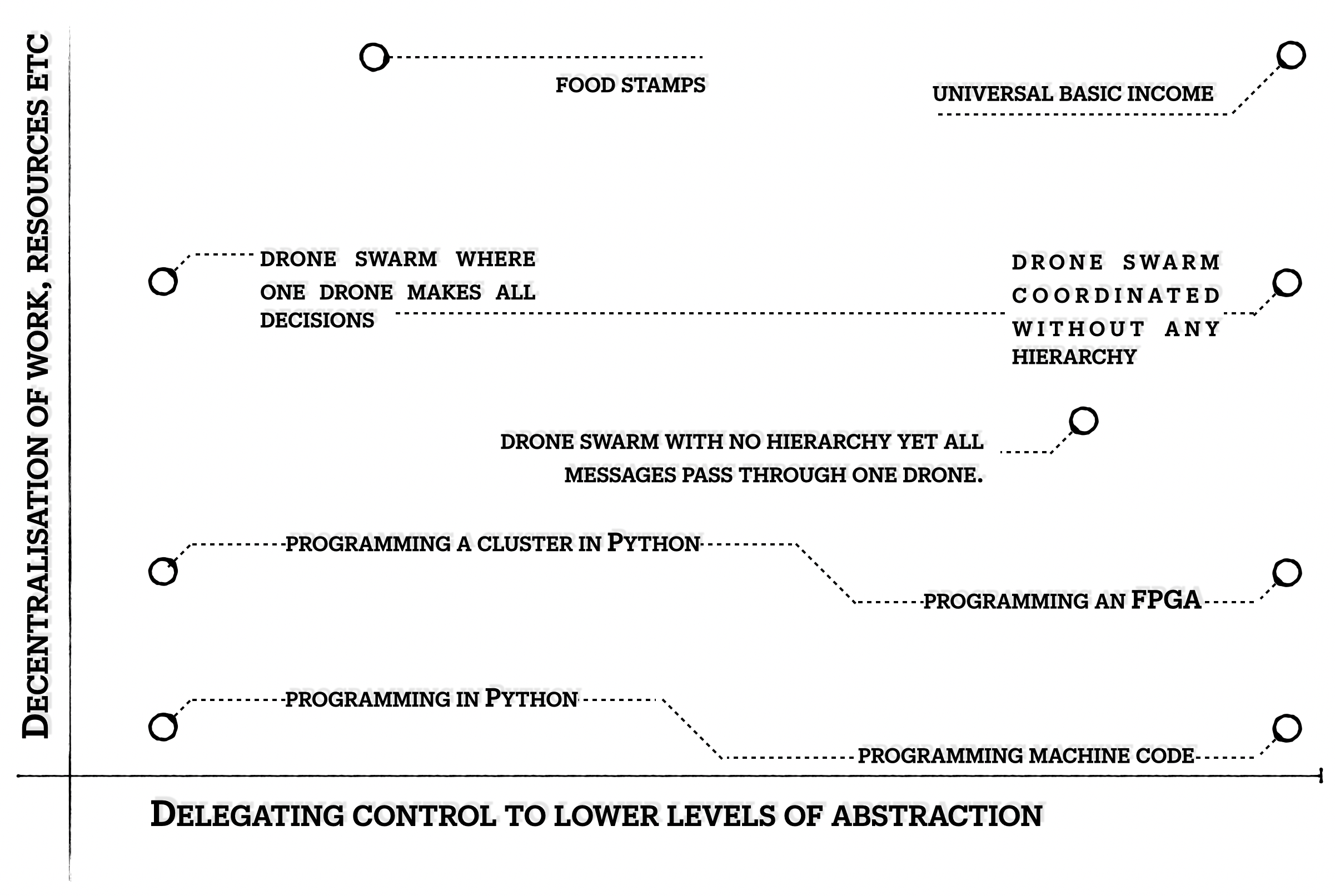}
\caption{Work, resources and so on can be distributed across a system. Distribution is distinct from delegation of control. For example, both food stamps and universal basic income might distribute the same amount of resources to the same people within an economic system, but the former more tightly constrains the recipients in how they use those resources (figure from \cite{bennett2025thesis}).}\label{fig3}
\end{figure}

Real systems occupy a continuum. Of course, there are limits on what sort of physical forms can store or process information in service of a goal \cite{sole2024}, and this may limit how much I can distribute work or delegate control. Delegating control often coincides with shifting to finer granularity and greater distribution. However, as computers illustrate, abstraction can also change without changing the spatial scale of components (Figures~\ref{fig2} and \ref{fig3}).

\subsection{Reasons to Suspect Delegation Facilitates Adaptation}  
Theorem~\ref{thm:stack} provides the formal claim. Here I briefly motivate why delegation should matter. Because perceived complexity is defined by the abstraction layer through which a system interacts with its environment \cite{bennett2024b}, measures of intelligence that ignore embodiment are incomplete. An ``objective'' measure must be enactive \cite{thompson2007}.

First, adaptation closer to the substrate can exploit more degrees of freedom and specialisation. Programming at a lower level of abstraction can yield large efficiency gains (e.g. C vs Python) \cite{oden2020}. Reconfigurable substrates such as FPGAs can push adaptation further down the stack, enabling even more specialisation and efficiency \cite{nguyen2020}.

Second, delegation can improve robustness via graceful degradation, avoiding single points of failure \cite{kuniavsky2010}. For example, to co-ordinate a drone swarm I could program one drone to plan the movements of the others. However, if that leader drone were damaged, the swarm would fail to co-ordinate. A more resilient approach is to program the swarm in a decentralised manner where each drone negotiates solutions with its neighbours, so the swarm has no single point of failure. Decentralised swarm control also emerges naturally, such as with flocks of starlings flying in complex formations \cite{cavagna2010}.

Third, in human organisations delegation shortens feedback loops. In military doctrine, ``mission command'' delegates decision-making to units closest to the situation, granting latitude to deviate from detailed orders in service of higher-level intent \cite{Shamir01102010}. This trades central control for responsiveness. See ESM Note~S7 for a concrete historical illustration (Nordbat~2 \cite{ingesson2016,ingesson2017}).

These are all intuition-building thought experiments rather than empirical proof, but they clarify the mechanism the theorem captures. To adapt at high levels, a system must retain enough slack at lower levels.

\section{What is Adaptation?}
I will discuss what it really means to learn and adapt, in terms of cause and effect. 
$$
\begin{tikzpicture}
    \node[state] (a) at (0,0) {cause};
    \node[state] (b) [right =of a]{effect};
    \path (a) edge (b);
\end{tikzpicture}
$$
To make accurate inferences a system must be able to tell the difference between what it has caused, and what it has merely observed \cite{merker2005,pearl2018,dawid2002}. It is often difficult to determine cause. `Confounding' occurs when an event appears to be the cause of the effect we're trying to predict, when in truth it is merely correlated. This can lead one to draw erroneous conclusions.
$$
\begin{tikzpicture}
    \node[state] (a) at (0,0) {not the cause};
    \node[state] (b) [right =of a] {effect};
    \node[state] (c) [above =of a] {cause};
    \path (c) edge (b);
    \path (c) edge (a);
    \draw[dotted] (a) -- (b);
\end{tikzpicture}
$$
By actively \textit{intervening} in the environment, a system can experiment to establish cause\footnote{Assuming such experimentation is enough to mitigate ambiguity and eliminate unknowns, and that a heuristic such as weakness is applied to ensure the system converges on variable definitions which represent cause accurately \cite{bennett2023c}.}. Then, it can construct a causal graph.
There are two ways to do that. One option would be to assume the nodes (variables, actions and other abstract objects and properties), and then learn the causal relations between them. This might work well in theory \cite{richens2024robust}, but it requires foreknowledge of objects and properties between which causal relations can be established. The alternative is to assume the causal relations (the edges in the graph), and then \textit{learn} the variables \cite{bennett2023c,bennett2025thesis,bennett2025e}. This needs only attraction and repulsion from physical states \cite{bennettwelshciaunica2024,bennett2025thesis}, otherwise known as valence. I take this latter approach. While causal discovery of this sort remains difficult to engineer in open-ended settings (e.g. unobserved confounders and combinatorial hypothesis spaces), I treat this as a modelling stance and focus on the stack-level delegation constraint formalised in Theorem~\ref{thm:stack}.

Learning nodes to fit a graph is like figuring out which computer program was used to generate a sequence. Assume I have fed a set of inputs into a computer. For each input the computer responded with a particular output. The computer runs a `policy' program which maps inputs to outputs\footnote{I call such programs policies, but this should not be confused with the more specific definition policy in reinforcement learning \cite{sutton2018}.}. That policy is the \textit{cause} of the input-output history observed so far. I want to know the policy that caused historical outputs so that I can use it to predict future outputs. The problem is that more than one policy could have produced the history we've observed, and each of those policies produce \textit{different} outputs when I give the computer inputs it hasn't encountered before. From a naive Bayesian perspective, all of them are equally likely to have caused the history I saw. If I wish to know which policy actually caused the outputs, I need a way to whittle down the remaining possibilities.

There are two ways to compare policies that look the same with respect to historical data, but differ with respect to possible future data \cite{bennett2024b}. One is to compare their \textit{form}, and choose policies which are \textit{simpler} in the sense of being more compressible. However, in an \textit{interactive} setting simplicity is an unreliable indicator of plausibility \cite{leike2015,bennett2024b}. The alternative is to compare policies by their \textit{function}, in terms of how they behave given inputs for which I do \textit{not} have historical data. Adaptive systems must satisfy functional \textit{constraints} \cite{sole2022}. I can compare systems by these constraints. Each policy implies a constraint on possible outputs given a set of possible inputs. The \textit{weaker} that constraint is, the more variety of future behaviour it permits. Put another way, if two policies satisfy a goal, then the \textit{weaker} policy satisfies it in more possible worlds\footnote{To the extent that such possible worlds can be differentiated from one another within the subjective constraint of an abstraction layer.}. Previous research showed that if two policies are identical with respect to historical data, then the policy most likely to have generated historical data is the \textit{weaker} \cite{bennett2024appendices,bennett2025thesis} (see ESM Supplementary Note~S3 for formal details.) of the two \cite{bennett2023c,bennett2025d,bennett2025e}. The policy most likely to have \textit{caused} behaviour is the \textit{weakest} \cite{bennett2023c}, and in experiments a system which preferred weaker policies outperformed one that preferred the simpler policies by $110-500\%$ \cite{bennett2023b}. This has been used to establish an upper bound on embodied `intelligence' \cite{bennett2024a}. I will use the term `w-maxing' to mean a system optimises for weak policies, and `simp-maxing' to mean a system optimises for simpler (minimum description length \cite{rissanen1978}) policies, to conform to the terminology used in previous work \cite{bennett2025b,bennett2025thesis,bennett2024appendices}. Put another way, if two hypotheses appear equally good with respect to past data, the w-maxing system will use policy weakness as a heuristic to decide which policy to use, whereas a simp-maxing system will use simplicity. 

Weakness is not the same thing as simplicity. Simplicity is a property of a policy's description in a chosen encoding. The weakness of a policy is the degrees of freedom remaining to a system that embodies that policy. Two correct policies can therefore have very different weakness even when their descriptions are equally short, or when one is syntactically longer. Weakness is also not mere truth set size. It counts how many distinct further commitments remain available to the agent given what it can express and implement at that abstraction layer. In this sense, w-maxing implements a least commitment heuristic stated at the level of function, rather than at the level of syntax. Since the completion structure depends on the vocabulary, weakness naturally incorporates embodiment. Changing sensors, effectors, or internal degrees of freedom changes $\mathfrak{v}$ and changes which refinements count as distinct. In the uniform case where all remaining completions are treated as equally plausible, maximising weakness coincides with a maximum entropy preference over viable futures.

An abstraction layer can be understood as a border between \textit{form} and \textit{function}. For example a program takes a \textit{form} when it is written in a language, but when it is run on a computer it has a particular \textit{function} or \textit{effect} upon the world. Simp-maxing is a matter of form, while w-maxing is a matter of function \cite{bennett2025b}.
When we account for interpretation between form and function it becomes clear that software is a physical thing, like a tool. A tool can serve more than one purpose. For example, a hammer is both a weapon and a paperweight. This is what ``weakness'' measures. A tool that implies a \textit{weaker} constraint can complete more tasks \cite{bennettmaruyama2022a} than another tool. By preferring weaker policies, one avoids unnecessary commitments, meaning the weakest policies complete a `wider range of tasks' than others (see ESM Note~S3 for discussion of weakness and related concepts, such as maximum entropy). To complete a `wider range of tasks' in the embodied sense, a tool must be both more sample and more energy efficient, and so the weakest policy is the most efficient in both senses. Put another way, the same physical matter can simultaneously contribute to many different tasks or computations superimposed over each other at different scales. A `weaker' policy serves more masters.

ML systems do not typically optimise for either weaker constraints on function (w-maxing), or simplicity of form (simp-maxing). This may explain some of their relative inefficiency \cite{bennett2022a,bennett2023b}. Even if they were w-maxing, the capacity for w-maxing is severely limited by the abstraction layer. An abstraction layer can be a ``causal isolator'' \cite{friston2023} between a software ``mind'' at higher levels of abstraction, and its environment at lower levels. Embodiment by another name. Different abstraction layers permit different upper bounds on policy weakness, in the same way that two bodies might differ in how able they are to achieve a given goal. Different embodiments give different upper bounds on `intelligence' \cite{bennett2024a}.

That biological self-organising systems adapt efficiently means they are able to embody weaker policies than today's ML systems. This capacity hinges on the abstraction layer \cite{bennett2024a}.
An abstraction layer is like a window through which only part of the world can be seen. It conveys an inductive bias that helps or hinders adaptation. When I optimise the system at a lower level of abstraction, this is like saying I can ``move'' this window to convey a different inductive bias. The bigger the window, the ``weaker'' the policy it is possible to learn. What this means is that intelligence is primarily a matter of the lens through which we interact with a problem. Hardware, not just software. In enactivist terms, an intelligent choice of abstraction layer is one that reduces the ``big world'' of all information to a ``small world'' of \textbf{relevant} information \cite{vervaeke2012,vervaeke2013a,vervaeke2013b}.
In a typical computer, lower levels of abstraction tend to be static. The window cannot move. I will show a static abstraction layer limits adaptability at higher levels of abstraction by limiting how weak the policies embodied by the system can be. Adaptability at high levels \textit{requires} adaptability at low levels, and delegation of control facilitates this.

\section{Formal Framework and Definitions}
The core claim and Theorem~\ref{thm:stack} depend upon the Stack Theory primitives defined below. The surrounding narrative provides intuition. The ESM contains additional motivation, proofs, and related technical material \cite{bennett2025thesis,bennett2023b,bennett2023c,bennett2024a,bennett2024b,bennettwelshciaunica2024,bennett2025e}.

Stack Theory describes an adaptive system in terms of the binary constraints or boundary conditions that define it. Embodiment is a choice of abstraction layer. This is a finite vocabulary of possible constraints (programs) representing the possible configurations of a body embedded in its environment \cite{bennett2025thesis}. These programs are
used to express “statements” constraining the body. Policies and tasks are defined as statements in this embodied language, and delegation is captured by how task constraints are arranged across layers. ESM Note~S5 provides a longer conceptual motivation and links to related perspectives.

For example, if a human raises their arm, then they are expressing a statement in their embodied language. That statement constrains what the body does next. The vocabulary $\mathfrak{v}$ of programs express this physical constraint. The inputs, outputs and policies of a system are all just subsets of its vocabulary. A $\mathfrak{v}$-task $\langle I, O\rangle$ is a set of inputs and corresponding outputs. If the $\mathfrak{v}$-task describes a `fit' organism, then the outputs it contains are `correct' according to natural selection. A $\mathfrak{v}$-task $\langle I,O\rangle$ can represent biological selection pressures by specifying which outputs are viable for which inputs. For example, an FEP-style agent can be treated as a task whose admissible futures minimise surprisal under an appropriate generative model \cite{ciaunica2023b,Friston2010}.

For readers who wish to check the formal dependencies (see also \cite{bennett2025thesis,bennett2024appendices}), the key definitions used in the theorem are summarised below.

\begin{enumerate}    \item \textbf{Environment, programs, vocabularies.} Fix a nonempty set $\Phi$ of mutually exclusive \textbf{states}. A (declarative) \textbf{program} is a set $p\subseteq \Phi$, and $p$ is \texttt{true} at $\phi\in\Phi$ iff $\phi\in p$. Let $P:=2^{\Phi}$ be the set of all programs. Embodiment is formalised as an \textbf{abstraction layer}, for which I single out a \emph{finite} \textbf{vocabulary} $\mathfrak{v}\subseteq P$.

    \emph{Intuition. A program is a set of states, and a vocabulary is the finite set of predicates the layer can express.}

    \item \textbf{Statements and truth sets.} The language generated by $\mathfrak{v}$ is
    \[
        L_\mathfrak{v}:=\{\; l\subseteq \mathfrak{v} : T(l)\neq \emptyset\;\},
        \qquad
        T(l):=\bigcap_{p\in l} p \subseteq \Phi,
    \]
    where $T(l)$ is the \textbf{truth set} (truth conditions) of the (conjunctive) statement $l$. By convention, $T(\emptyset)=\Phi$.
    A statement $l$ is \texttt{true} at state $\phi$ iff $\phi\in T(l)$.

    \emph{Intuition. A statement is a conjunction of vocabulary programs. Its truth set is the states where all of them hold.}

    \item \textbf{Completions, extensions, weakness.} A \textbf{completion} of a statement $l$ is any statement $l'\in L_\mathfrak{v}$ such that $l\subseteq l'$. The \textbf{extension} of $l$ is the set of all its completions.
    \[
        \Ext{l} := \{\;l'\in L_\mathfrak{v} : l\subseteq l'\;\}.
    \]
    For a set $X\subseteq L_\mathfrak{v}$, define $\Ext{X}:=\bigcup_{x\in X}\Ext{x}$. (Throughout, $\Ext{\,\cdot\,}$ is always taken in the ambient language $L_\mathfrak{v}$ determined by context.)
    The \textbf{weakness} of a policy $\pi$ is the cardinality of its extension $|\Ext{\pi}|$.

    \emph{Intuition. A completion adds further constraints while staying compatible with what has already been fixed. Weakness counts how many distinct completions remain available in the agent's own embodied language. It is therefore a measure of optionality or autonomy at that abstraction layer. Weakness is not a syntactic notion like description length. A policy can be very short yet highly committal, or long yet weak, depending on how the vocabulary factors the world. Changing embodiment changes the vocabulary and therefore changes which refinements count, which is intended.}

    \item \textbf{Tasks and hierarchy.} A $\mathfrak{v}$-\textbf{task} $\alpha$ is a pair $\langle I_\alpha, O_\alpha\rangle$ where $I_\alpha\subseteq L_\mathfrak{v}$ is a set of admissible \textbf{inputs} and $O_\alpha\subseteq \Ext{I_\alpha}$ is a set of \textbf{correct outputs}. Tasks form a hierarchy. If $\alpha\sqsubset \omega$ then $\omega$ is a \textbf{parent} of $\alpha$, meaning $I_\alpha\subseteq I_\omega$ and $O_\alpha\subseteq O_\omega$.

    \emph{Intuition. A task specifies which outputs count as correct for each input. The hierarchy orders tasks by adding constraints.}

    \item \textbf{Policies, correctness, inference, learning.} A \textbf{policy} is a statement $\pi\in L_\mathfrak{v}$. A policy is \textbf{correct} for $\alpha$ iff
    \[
        \pi \in \Pi_\alpha := \{\pi \in L_\mathfrak{v} : \Ext{I_\alpha} \cap \Ext{\pi} = O_\alpha\}.
    \]
    Given input $i\in I_\alpha$ and a policy $\pi$, to \textbf{infer} is to choose $o \in \Ext{i}\cap \Ext{\pi}$. A task is complete if $o\in O_\alpha$. Given $\alpha\sqsubset\omega$, to \textbf{learn} is to choose a policy $\pi\in\Pi_\alpha$, and $\omega$ has been \textbf{learned} from $\alpha$ if one chooses $\pi\in\Pi_\omega$.

    \emph{Intuition. A policy is a statement that filters which outputs are allowed. Learning chooses a policy whose completions match the observed input output pairs.}

    \item If $\alpha_1 \sqsubset \alpha_2 \sqsubset \omega$ then $\alpha_1$ contains fewer examples of $\omega$ than $\alpha_2$. System $\mathfrak{a}$ is more intelligent than $\mathfrak{b}$ if it learns $\omega$ from fewer examples. For example $\mathfrak{a}$ is more intelligent than $\mathfrak{b}$ if $\mathfrak{a}$ learns $\omega$ from either $\alpha_1$ or $\alpha_2$, while $\mathfrak{b}$ learns $\omega$ from $\alpha_2$ but not from $\alpha_1$.

    \emph{Intuition. The point of this example is that tighter constraints at one layer limit how weak a compatible higher-layer policy can be. This is why delegation matters for efficiency.}

    \item A \textbf{causal identity} for $\mathfrak{b}$ is a policy classifying the interventions undertaken by $\mathfrak{b}$.

    \emph{Intuition. Collective identity is the existence of at least one policy that is simultaneously viable for the parts and for the whole.}

    \item $\Gamma_P$ contains every task in every vocabulary. For every $\rho \in \Gamma_P$ there exists an \textbf{uninstantiated-task} function $\lambda_\rho : 2^P \rightarrow \Gamma_P.$
    $\lambda_\rho$ takes a vocabulary $\mathfrak{v}' \in 2^P$ and returns the highest-level child of $\rho$ that is a $\mathfrak{v}'$-task.

    \emph{Intuition. An uninstantiated task is a task template that can be applied to different vocabularies. This lets the same selection pressure be expressed across layers.}

    \item The utility of a $\mathfrak{v}$-task $\gamma$ is $\epsilon(\gamma) = \underset{\pi \in \Pi_{\gamma}}{\max} \left(\lvert \Ext{\pi} \rvert - \lvert O_{\gamma} \rvert \right).$

    \emph{Intuition. Utility increases when the correct-output set can be realised by many policies. This captures adaptability under the embodied language.}

\end{enumerate}

I set out to compare computers with biological systems. Both are stacks of abstraction layers. Hence I define a stack (meaning a multilayer architecture or MLA) below. I do this to describe biological and contemporary computer systems in comparable terms. 

{\small
\begin{itemize}
    \item The \textbf{stack} of layers of increasing abstraction is a sequence of uninstantiated tasks $\langle \lambda^0, \lambda^1, \dots, \lambda^n \rangle$ such that $\lambda^{i+1} \sqsubset \lambda^{i}$. Lower in the stack is a lower level of abstraction.

    \item Given a vocabulary $\mathfrak{v}\subseteq P$ and a policy $\pi\in L_\mathfrak{v}$, the induced \textbf{abstractor} is
    \[
        \mathfrak{f}(\mathfrak{v},\pi) := \{\,T(o) : o \in \Ext{\pi}\,\} \subseteq P.
    \]
    That is, the next-layer vocabulary consists of the truth sets of completions compatible with $\pi$.

    \item The \textbf{state} of the stack (MLA) is a sequence of policies $\langle \pi^0, \pi^1, \dots, \pi^{n} \rangle$ and a sequence of vocabularies $\langle \mathfrak{v}^0, \mathfrak{v}^1, \dots, \mathfrak{v}^n \rangle$ such that for each $i<n$,
    \[
        \mathfrak{v}^{i+1} = \mathfrak{f}(\mathfrak{v}^i, \pi^i)
        \quad\text{and}\quad
        \pi^i \in \Pi_{\lambda^{i}(\mathfrak{v}^{i})}.
    \]
\end{itemize}
\Intuition The stack records tasks at increasing abstraction, and each layer is constrained by the layer below. The abstractor tells you which higher-level predicates become available once a lower-level policy has fixed which completions are possible. The state records the current policy and induced vocabulary at each layer.

}

\noindent I say the stack (MLA) is \textbf{over-constrained} when there exists $i < n$ such that $\Pi_{\lambda^{i}(\mathfrak{v}^{i})} = \emptyset$, and \textbf{multilayer-causal-learning} (MCL) is when the stack is not over-constrained. Over-constraint means some layer cannot realise any policy compatible with the higher-level task. The MCL is the boundary where each layer still has at least one viable policy, so collective identity is possible.

\section{Delegated Control and Collective Identity}
The vocabulary $\mathfrak{v}$ of an abstraction layer can describe a distributed system. For example, it could describe an organ $\omega$ formed of two cells $\alpha_1$ and $\alpha_2$. The behaviour of two cells as individuals is formalised as two $\mathfrak{v}$-tasks $\alpha_1$ and $\alpha_2$. The collective behaviour of the two cells is a third $\mathfrak{v}$-task $\omega$ s.t. $\alpha_1 \sqsubset \omega$ and $\alpha_2 \sqsubset \omega$. Intuitively, this just means the two cells are parts of a larger whole. \textit{If} there exists a correct policy for $\omega$, then the cells \textit{can} behave in service of a common goal and thus have a common identity (for example if both cells use policies in $\Pi_\omega = \Pi_{\alpha_1} \cap \Pi_{\alpha_2}$). 
$$
\begin{tikzpicture}
    \node[state] (a) at (0,.75) {$\alpha_1$};
    \node[state] (b) at (0,0) {$\alpha_2$};
    \node[state] (c) at (3,.375){$\omega$};
    \path (a) edge (c);
    \path (b) edge (c);
\end{tikzpicture}
$$
The probability that the parts of a system embody a collective policy is higher if they embody weaker policies.

Classical cybernetics characterises regulation in terms of \emph{variety}. Ashby's law of requisite variety says that a regulator must possess at least as much variety as the disturbances it must absorb. Delegated control can be read as a constructive way to generate such requisite variety by distributing adaptive degrees of freedom across layers rather than concentrating them in a single controller \cite{ashby1957,wiener1948cybernetics}. This also connects to cybernetic accounts of hierarchical viability and self-regulation, such as Beer's Viable System Model \cite{beer1995brain}. This is because weaker policies are more likely to generalise between $\mathfrak{v}$-tasks \cite{bennett2025e}. 

Consider the aforementioned $\alpha_1$ and $\alpha_2$. A sufficiently weak policy at the level of individual cells is necessary for the collective to have an identity, because only the weakest policies in $\Pi_{\alpha_1}$ and $\Pi_{\alpha_2}$ will also be in $\Pi_\omega = \Pi_{\alpha_1} \cap \Pi_{\alpha_2}$. 
A collective of cells remains cells regardless of whether I deem it to be an organ. The parts are still there. In terms of the stack, the cells $\alpha_1$ and $\alpha_2$ and the collective $\omega$ share a vocabulary. Just as the outputs of $\alpha_1$ can affect the inputs of $\alpha_2$ and vice versa, the outputs of $\omega$ determine the inputs of $\alpha_1$ and $\alpha_2$. The difference is a matter of scale. Just as w-maxing facilitates causal learning at the scale of an individual \cite{bennett2023c}, in the context of an MCA it will facilitate learning at multiple scales, because the system as a whole is just another $\mathfrak{v}$-task. 

Mechanistically, ``negotiation'' between levels is represented as constraint compatibility between policy sets. Individual components select policies that remain correct for their local tasks while being consistent with the collective task, and the collective task in turn constrains components by shaping the inputs they experience. When the intersection of viable policies becomes empty, no collective policy exists and collective identity collapses (a failure mode analysed in Section~\ref{sectionCancer}). In hybrids, the designer mediates this negotiation by adjusting boundary conditions (which reshape the component policy sets) rather than by directly specifying a single low-level policy.

\subsection{The Stack}

From the enactive perspective, a human is something cells \textit{do}, rather than \textit{are}. 
An organ boils down to a collection of cells. The cells are an abstraction layer in which the collective policy of the organ is expressed, like software. However, software is just a \textit{state} of hardware \cite{bennett2024a}. Put another way, `organ' is a label I \textit{ascribe} to the behaviour of some cells, and `human' is a label I ascribe to the behaviour of some organs. `x86' is a label I ascribe to a \textit{product} of human behaviour. It is a static abstraction layer an engineer has embodied in silicon and then left alone. `fire and forget'. Humans act as the lower layer from which x86 emerges, but once it is embodied in silicon we disconnect it from our stack. We causally `intervene' in the stack. To understand the effect of this intervention, I need to abstract away the lower levels. I must ``abstract away'' cells and focus on the consequences of their \textit{collective} behaviour. 

The cells are tasks. For tasks at higher levels of abstraction, capacity for w-maxing depends on lower levels. A \textbf{policy} $\pi \in \Pi_{\alpha_1} \cap \Pi_{\alpha_2}$ \textbf{implies an abstraction layer} in the sense that for every $o \in \Ext{\pi}$ there exists $p\in P$ such that $T(o)=p$ (i.e.\ $p$ and $o$ are \texttt{true} of exactly the same states). Hence I can transform policies to abstraction layers, using \textbf{a function} $\mathfrak{f}$ that \textbf{takes the extension} $\Ext{\pi}$ of a collective's policy $\pi$, and \textbf{returns a new vocabulary} $\mathfrak{v'} = \{\,T(o): o \in \Ext{\pi}\,\}$. I can then use a sequence of uninstantiated tasks $\langle \lambda^0, \lambda^1 ... \lambda^n \rangle$ s.t $\lambda^{i+1} \sqsubset \lambda^{i}$ to represent selection pressures that define `correctness' at every level. At each successive level, `correct' gets more specific. Because everything that happens in a higher level is in some sense correct according to lower levels, each higher level must more tightly constrain what is considered to be correct behaviour if that higher level is to exist in a meaningful sense. 

$$
\begin{tikzpicture}
    \node[state] (a) at (0,0) {$\lambda^{i-1}$};
    \node[state] (b) [right =of a] {$\lambda^{i}$};
    \node[state] (c) [right =of b] {$\lambda^{i+1}$};
    \path (a) edge (b);
    \path (b) edge (c);
\end{tikzpicture}
$$

To construct layers I would start with a vocabulary $\mathfrak{v}^i$ at $i=0$, for example implied by a collective of cells or a lower level system. If this system exhibits goal-directed behaviour, then it is given by the $\mathfrak{v}^i$-task $\lambda^i(\mathfrak{v}^i)$ and a policy $\pi^i \in \Pi_{\lambda^{i}(\mathfrak{v}^i)}$. $\pi^i$ implies behaviour. I translate policy into an abstraction layer using the aforementioned $\mathfrak{f}$. $\mathfrak{f}$ is applied to $\Ext{\pi^i}$ to obtain a vocabulary $\mathfrak{v}^{i+1}$, and then $\lambda^{i+1}$ is applied to $\mathfrak{v}^{i+1}$ to obtain a task $\lambda^{i+1}(\mathfrak{v^{i+1}})$. If $\Pi_{\lambda^{i+1}(\mathfrak{v^{i+1}})} \neq \emptyset$, then there exists a common purpose or ``self'' at this level of abstraction and I select another policy $\pi^{i+1} \in \Pi_{\lambda^{i+1}(\mathfrak{v^{i+1}})}$. 

In narrative terms, adaptation at higher levels depends on adaptation at lower levels. I now state \emph{The Law of the Stack} and give a brief proof sketch (the ESM provides a full proof and additional lemmas).

\noindent\textit{Formally the next theorem quantifies a delegation bottleneck, that utility at layer $i{+}1$ is upper-bounded by the weakness (slack) of the realised policy at layer $i$.}

\begin{theorem}[The Law of the Stack]\label{thm:stack}Let $\langle \lambda_0,\lambda_1,\dots,\lambda_n\rangle$ be a stack (MLA) with abstractor
$\mathfrak{f}$ and state $\langle (\pi_0,\dots,\pi_n),(v_0,\dots,v_n)\rangle$ such that
$v_{i+1}=\mathfrak{f}(v_i,\pi_i)$ and $\pi_i\in \Pi_{\lambda_i(v_i)}$ for all $i<n$.
Fix any $i<n$, and write
\[
v_{i+1}=\mathfrak{f}(v_i,\pi_i),\qquad \gamma_{i+1}=\lambda_{i+1}(v_{i+1}).
\]
Assume $\Pi_{\gamma_{i+1}}\neq\emptyset$ so that $\epsilon(\gamma_{i+1})$ is well-defined.
Then the achievable utility at layer $i\!+\!1$ is upper-bounded by the weakness of
the policy at layer $i$ in the following sense.
\[
\epsilon(\gamma_{i+1}) + |O_{\gamma_{i+1}}| \;\le\; 2^{|\Ext{\pi_i}|}.
\]
Equivalently,
\[
|\Ext{\pi_i}| \;\ge\; \log_2\!\big(\epsilon(\gamma_{i+1})+|O_{\gamma_{i+1}}|\big).
\]
In particular, increasing $\epsilon(\lambda_{i+1}(v_{i+1}))$ requires (at least)
increasing $|\Ext{\pi_i}|$, i.e. weakening the policy at the lower layer.
\end{theorem}
\Intuition The bound says that higher-layer utility cannot grow without leaving enough slack in the realised policy one layer down. A narrow lower-layer policy induces a small next-layer vocabulary and this caps achievable outputs.

\begin{proof}[Proof sketch] Since $\Pi_{\gamma_{i+1}}\neq\emptyset$, utility $\epsilon(\gamma_{i+1})$ is attained by some correct policy at layer $i\!+\!1$,
whose extension is a subset of $L_{v_{i+1}}$, so $\epsilon(\gamma_{i+1})+|O_{\gamma_{i+1}}|
\le |L_{v_{i+1}}|$. Because $v_{i+1}$ is finite, $|L_{v_{i+1}}|\le 2^{|v_{i+1}|}$.
By construction of the abstractor, $v_{i+1}$ is the image of $\Ext{\pi_i}$ under the
map $o\mapsto T(o)$, hence $|v_{i+1}|\le |\Ext{\pi_i}|$. Combining yields
$\epsilon(\gamma_{i+1})+|O_{\gamma_{i+1}}|\le 2^{|\Ext{\pi_i}|}$.
\end{proof}
\ProofIntuition The proof sketch counts how many distinct higher-layer behaviours exist given the induced vocabulary at layer $i{+}1$. The induced vocabulary is bounded by the number of completions left open by the realised policy at layer $i$.

\subsection{Delegation and variational free energy}
The Law of the Stack is a structural constraint. It says that degrees of freedom in higher layers is bounded by lower-layer slack. This aligns with the Free Energy Principle (FEP), which frames adaptive behaviour as variational free energy minimisation \cite{Friston2010,friston2013,parr2024}. In the ESM (``Viability, lived history, and variational free energy'') I show that under a uniform viability prior, minimising free energy is equivalent to maximising the entropy (log-count) of viable future continuations. In Stack Theory terms, this is exactly w-maxing.

Combining that equivalence with Theorem~\ref{thm:stack} yields the corollary that reducing free energy at an abstract layer requires sufficient delegation (sufficiently weak realised policies) at lower layers. Intuitively, if lower layers are over-constrained, the induced higher-layer language is small, so even a maximum-entropy posterior cannot keep many viable futures open.

\noindent\textit{The next corollary makes the delegation dependence formal. Under the uniform viability prior, abstract-layer free energy cannot be reduced below a floor set by lower-layer slack.}\\

\begin{corollary}[Delegation is necessary for lowering free energy]\label{cor:fep_delegation}
Fix a planning horizon and let $\mu$ denote the viability task at layer $i{+}1$.
Write $O_\mu\subseteq L_{v_{i+1}}$ for the set of viable future continuations at that horizon.
For any policy $\pi_{i+1}$ at layer $i{+}1$ with $\Omega_{\pi_{i+1}}\neq\emptyset$, let $\Omega_{\pi_{i+1}} = \Ext{\pi_{i+1}}\cap O_\mu$ be the set of viable futures compatible with $\pi_{i+1}$, and let $q_{\pi_{i+1}}$ be the uniform distribution on $\Omega_{\pi_{i+1}}$.
Under the uniform viability prior assumed in the ESM, the corresponding (base-2) free energy satisfies
\[
F_2(q_{\pi_{i+1}})=\log_2|O_\mu|-\log_2|\Omega_{\pi_{i+1}}|.
\]
Moreover, $\Omega_{\pi_{i+1}}\subseteq L_{v_{i+1}}$, hence $|\Omega_{\pi_{i+1}}|\le |L_{v_{i+1}}|\le 2^{|\Ext{\pi_i}|}$ as in the proof of Theorem~\ref{thm:stack}, where $\pi_i$ is the realised policy at layer $i$ that induces $v_{i+1}$.
Therefore,
\[
F_2(q_{\pi_{i+1}})\;\ge\;\log_2|O_\mu|-|\Ext{\pi_i}|.
\]
In particular, free energy minimisation at layer $i{+}1$ is bottlenecked by $|\Ext{\pi_i}|$, so reducing $F_2(q_{\pi_{i+1}})$ requires sufficient lower-layer slack, i.e.\ a weak enough realised policy at the lower layer.
\end{corollary}
\Intuition In active inference terms, free energy can only be reduced by keeping enough viable futures in play (high entropy). The Law of the Stack proves that if lower layers are frozen or over-constrained, there is an entropy bottleneck, so abstract-layer free energy has a floor.\\

\noindent\textbf{Remark (accuracy and complexity).} In general, variational free energy admits a decomposition into an \emph{accuracy} term and a \emph{complexity} term. The uniform-viability model used in Corollary~\ref{cor:fep_delegation} folds accuracy into hard constraints, namely the viability constraint $\omega\in\Omega$, and the history-consistency restriction $\pi\in\Pi_{\mathfrak{h}}$ within the confines of an embodied vocabulary $\mathfrak{v}$. Utility and viability constrain what counts as accurate behaviour, and delegation affects residual flexibility \emph{within} that accuracy envelope. \\

Intervening in the stack limits the extent to which control can be delegated, and thus limits capacity for w-maxing or MCL. It makes lower layers of abstraction \textit{static} in the sense that they no longer change to facilitate adaptation at higher levels. By Theorem~\ref{thm:stack}, any increase in achievable utility at layer $i{+}1$ requires (at least) an increase in the extension size of the realised policy at layer $i$, i.e. support for weaker constraints below.

\subsection{Human Abstractions}
Humans are not just part of MCAs. A human \textit{is} an MCA. Human abstractions are emergent behaviours of that MCA.
$$
\begin{tikzpicture}
    \node[state] (a) at (0,0) {$cells$};
    \node[state] (c) [right =of a] {$humans$};
    \node[state] (d) [right =of c] {$abstractions$};
    \path (a) edge (c);
    \path (c) edge (d);
\end{tikzpicture}
$$
When we humans create a computer, we create a machine that embodies a subset of our more abstract behaviours. An engineer hard-wires those behaviours into the instruction set architecture (ISA), and we all then ascribe meaning to that behaviour. Two results can look the same at the higher level (e.g. the number two, spoken by a human or displayed on a screen), but be fundamentally different at the lower level (e.g. resources consumed, data format etc). The results are only the same if one ignores the lower levels of abstraction.

Software might be agentic, but computer hardware is passive. Control is top-down. Once I have built an ISA, it exists in the environment independently of the human MCA from which it emerged. It is static. It cannot adapt, which is why there is demand for field programmable gate array (FPGA) (a reconfigurable processor substrate). 
$$
\begin{tikzpicture}
    \node[state] (b) at (0,0) {$humans$};
    \node[state] (c) [right =of b] {$ISA$};
    \node[state] (d) [right =of c] {$abstractions$};
    \path (b) edge (c);
    \path (c) edge (d);
\end{tikzpicture}
$$
If an abstraction layer is a window onto the world, then \textit{utility} is how ``good'' the view is for a task. A static abstraction layer is a window that cannot move, while an MCA can change policy at lower levels, to move the window into an ideal location. 
When I create an ISA I insert a static abstraction layer at an intermediate level, so the vocabulary at the highest levels is limited to what I can do with those abstractions I have embodied in the ISA. In comparison, an MCA can learn policies at lower levels, to change the vocabulary of higher levels and facilitate \textit{multilayer} causal learning. This shows how adaptability at higher levels demands adaptability at lower levels. I am not suggesting lower levels \textit{only} optimise for utility at higher levels. After all, an increase in utility may increase the computational cost of searching for correct policies. Furthermore, a spatially extended environment constrains the size of vocabularies at all levels \cite{bennett2024b}, because bounded systems can contain only finite information \cite{bekenstein1981}. I mean lower levels learn an inductive bias for higher levels, balancing utility with other factors in service of goals at every scale. 

\section{Cancer-Like Failure Modes} \label{sectionCancer}

I have explored some advantages of bio-mimetic distribution and delegation. Are there disadvantages? Yes. A state analogous to cancer exists for distributed systems with insufficient delegation in adverse conditions. Some have argued cancer is a loss of collective identity \cite{levin2021,davies2011}. In a biological system, when cells are isolated from the informational structure of their collective, they lose their high level identity. When that happens, they independently pursue their own lower level goals. They revert to primitive transcriptional behaviour, while the collective of which they were formerly part continues on without them. They escape from immune control mechanisms by becoming \textit{less} differentiated, and closer to embryonic states. 

In the context of my formalism, a state analogous to cancer occurs when a system is `over-constrained' so that no collective policy exists, while the capacity for independent action remains. For example, assume a collective represented by a task. Adverse environmental conditions are represented as a task for which fewer correct policies exist than less adverse environmental conditions. Likewise excessive top down control can limit adaptability, and thus the number of correct policies which exist. If the constraints are too tight, then no correct policies will exist and the collective's identity will be lost.

\subsection*{Formal characterisation of the cancer-analogue condition}

\noindent\textit{Formally, the next definition and proposition show that if boundary conditions over-constrain a distributed system so that no collective policy exists, then any realised behaviour must be splintered, because it can only satisfy a proper subset of parts.}\\

This is a formal analogue rather than a biological claim. It captures the failure mode where the collective-policy intersection disappears while local viability persists, so continued function is only possible via coherent subcoalitions. In Stack Theory terms, this is the zero-slack limit of the delegation bottleneck quantified by Theorem~\ref{thm:stack}. Tightening lower-layer constraints can shrink the induced higher-layer feasible set all the way to emptiness.

To make the above precise, let $\alpha_1,\dots,\alpha_m$ be $v$-tasks representing parts of a distributed system, with policy sets $\Pi_{\alpha_j}\subseteq L_v$, and let the collective task be $\omega$ with collective-policy set $\Pi_\omega := \bigcap_{j=1}^m \Pi_{\alpha_j}$ (collective identity at $v$ requires $\Pi_\omega\neq\emptyset$).

\begin{definition}[Restricted policies, over-constraint, and splinters]\label{def:splinter}
To model top-down boundary conditions, fix restricted policy sets $\widehat{\Pi}_{\alpha_j}\subseteq \Pi_{\alpha_j}$ and define the restricted collective set $\widehat{\Pi}_\omega := \bigcap_{j=1}^m \widehat{\Pi}_{\alpha_j}$.
We call the system \textbf{over-constrained at $v$} (under the boundary conditions) if $\widehat{\Pi}_\omega=\emptyset$.
A \textbf{splinter} is a non-empty index set $S\subseteq\{1,\dots,m\}$ such that $\bigcap_{j\in S}\widehat{\Pi}_{\alpha_j}\neq\emptyset$.
\end{definition}

\begin{proposition}[Over-constraint implies loss of collective policy; any realised behaviour is splintered]\label{prop:splinter}
Assume at least one part remains locally viable, i.e.\ $\widehat{\Pi}_{\alpha_j}\neq\emptyset$ for some $j$.
If the system is over-constrained at $v$ (so $\widehat{\Pi}_\omega=\emptyset$), then there is no policy simultaneously feasible for all parts. Moreover, for any realised policy $\pi\in L_v$, the set of parts for which $\pi$ remains feasible,
\[
S(\pi):=\{\,j\in\{1,\dots,m\}:\pi\in \widehat{\Pi}_{\alpha_j}\,\},
\]
is a proper subset of $\{1,\dots,m\}$. In particular, any continued function can only proceed via a splinter (Definition~\ref{def:splinter}).
\end{proposition}

\par\noindent\textbf{Proof. }
If some $\pi$ were feasible for all parts then $\pi\in \bigcap_{j=1}^m \widehat{\Pi}_{\alpha_j}=\widehat{\Pi}_\omega$, contradicting $\widehat{\Pi}_\omega=\emptyset$. Hence every realised policy fails at least one part's constraints, so $S(\pi)\neq\{1,\dots,m\}$ for all $\pi\in L_v$. If $S(\pi)\neq\emptyset$ then $\pi\in\bigcap_{j\in S(\pi)}\widehat{\Pi}_{\alpha_j}$, so $S(\pi)$ is a splinter. If $\widehat{\Pi}_{\alpha_j}\neq\emptyset$ then $\{j\}$ is a splinter since $\bigcap_{k\in\{j\}}\widehat{\Pi}_{\alpha_k}=\widehat{\Pi}_{\alpha_j}\neq\emptyset$. \hfill $\square$\\

\noindent\textbf{Remark (integration and $\Phi_{\mathrm{IIT}}$).} Proposition~\ref{prop:splinter} characterises collective identity as the existence of a non-empty intersection of feasible policy-sets. This is a structural notion of integration. In the Stack Theoretic depiction of consciousness, a loss of collective identity would necessarily imply loss of selves (e.g. reafference) and the disintegration of phenomenal content \cite{bennett2025thesis,bennettwelshciaunica2024,bennett2026a}. Beyond Stack Theory, Integrated Information Theory formalises integration via the quantity $\Phi_{\mathrm{IIT}}$, which decreases when a system decomposes into weakly coupled parts \cite{tononi2004,oizumi2014}. Under that lens, splintering can be read as a loss of integration, rather than merely an over-constraint at one layer.\\

\noindent\textbf{Remark (temporal splintering and negotiation).} The set-intersection definition in Proposition~\ref{prop:splinter} is static. Hybrid agents often face transient conflicts where $\widehat{\Pi}_{\omega}(t)=\emptyset$ for short intervals. A natural dynamic refinement is to treat each part's feasible set as time-indexed and to model resolution as a higher-level regulatory process that reallocates variety across layers to restore a non-empty viable intersection. This is closely aligned with cybernetic treatments of adaptive organisation, including Beer's Viable System Model and Ashby's requisite variety principle \cite{beer1995brain,ashby1957}. Future work may further extend this framing of cancer using the time semantics module associated with Stack Theory \cite{bennett2026a}.\\

As task difficulty increases (or top-down control tightens), the set of correct collective policies shrinks. A task $\alpha$ is easier than $\omega$ if $|\Pi_\alpha|>|\Pi_\omega|$. In Stack Theory terms, a cancer-like breakdown occurs when a policy change at layer $i{-}1$ induces a new collective task at layer $i$ with \emph{no} correct policies, $\Pi_{\lambda^i(v^i)}=\emptyset$, while components retain the capacity for independent action.

When $\Pi_{\lambda^i(v^i)}=\emptyset$, the collective has no viable way to act as a coherent unit. Any continued function therefore requires dropping constraints, parent task splintering into ever narrower child tasks until those tasks are narrow enough to admit a policy. Part of the collective stops participating in the collective policy and reverts to a lower-level policy set with a much larger extension (more autonomy). This is a form of ``selective forgetting'' \cite{bennett2023c} applied to constraints rather than data. 

This is analogous to cancer. $\alpha$ is isolated and can now pursue goals which do not align with those of the collective. This suggests systems should be minimally constrained so as to ensure graceful degradation or `sloppy fitness' \cite{agosta2010}, maintaining core functionality in as wide a variety of unanticipated circumstances as possible. The more difficult the environment or the more top-down the system is controlled, the more tightly constrained behaviour becomes until the system becomes over-constrained and splinters. Co-operation becomes impossible and the parts of a system start to compete. Conversely, if control of a system is delegated to the lowest levels, then adverse conditions are less likely to eliminate all correct policies. Put another way, there will be less incentive for the parts of a system to splinter, as that would not greatly increase the number of correct policies available to them. A state analogous to cancer occurs in a distributed system when there is an over-reliance on top-down causation to make the parts of a system conform to a collective identity they are ill suited to, under the circumstances.

\section{Hybrid Creation as Weak Boundary-Condition Design}
Consider the aforementioned example of the organoid adapted to play the game Pong with notable sample efficiency relative to conventional reinforcement-learning baselines \cite{khajehnejad2024}. Hybrid systems like organoids are not fixed interpreters of software-level policy, but collectives whose lower-level parts retain bottom-up agency. In such systems, \emph{hybrid creation} is the act of designing an interface and environment that reshapes the feasible policy space of the biological components. Rather than ``programming'' the parts directly, the designer engineers boundary conditions so that the remaining viable collective policies implement a target input-output relation at the interface. Practically, hybrid creation is iterative: \begin{enumerate}
    \item choose boundary conditions, then
    \item observe emergent collective behaviour, then
    \item tighten or relax constraints to steer the intersection of viable policy sets toward the target interface behaviour and
    \item repeat.
\end{enumerate}

Similar boundary-condition design issues arise in other bio-hybrid constructs (e.g.\ xenobot-like cell collectives) and in soft-robotic substrates where not just software but morphology and material properties shape the feasible low-level policy space.

The designer selects boundary conditions that reshape each component's viable policy set, hoping that the intersection admits a collective policy that realises the desired input-output mapping at the interface. 

According to The Law of the Stack, achieving higher utility at an interface layer requires sufficiently weak policies at lower layers. Therefore, boundary conditions should be as \emph{weak} (least exclusionary) as possible while still making the target behaviour achievable and preserving a non-empty intersection of component policy sets (collective identity). Overly tight constraints can collapse that intersection and force failure, while overly loose constraints may fail to stabilise any desired collective policy. A schematic formalisation of this boundary-condition design view is provided in the ESM.

\section{Conclusions and Future Research} \label{sectionConclusion}
Biological systems often adapt efficiently because they delegate adaptive control across scales. Using Stack Theory \cite{bennett2025thesis}, I formalised delegation as the depth to which adaptation is permitted in a stack of abstraction layers and proved \emph{The Law of the Stack} (Theorem~\ref{thm:stack}). Utility one layer up is bounded by the weakness of the realised policy one layer down. Delegation is a structural prerequisite for high-level adaptability.

This constraint is compatible with active inference. Under a simple viability prior, minimising variational free energy is equivalent to maximising entropy over viable futures (w-maxing; see ESM ``Viability, lived history, and variational free energy''). Corollary~\ref{cor:fep_delegation} shows that free energy minimisation inherits the same bottleneck, which is that abstract-layer free energy cannot be reduced without sufficient lower-layer slack.

The implication for AI is not that computation cannot be intelligent, but that contemporary AI stacks tend to concentrate adaptation at high software layers while freezing lower layers, thereby throttling achievable weakness and adaptation efficiency. For hybrid agents (organoids, soft robots, and human--AI organisations) the design problem is to impose boundary conditions, constraining parts just enough to elicit target collective behaviour while preserving a non-empty intersection of viable policies (collective identity). When that intersection collapses, the system can fragment into competing subagents (Section~\ref{sectionCancer}).

Future work should test these predictions by varying which layers are permitted to adapt (e.g.\ hardware-in-the-loop learning, morphologically adaptive substrates, and biologically integrated controllers) and measuring resulting degrees of freedom, free energy, and robustness. More broadly, this paper is a stepping stone toward a unified account of delegation, weakness/max-entropy heuristics, and multiscale agency in the larger Stack Theory programme \cite{bennett2025thesis}.

To speculate about future research beyond experimental verification, \emph{Stack Theory} is a lens that can be applied generally. For example, a number of software-first approaches to AI already approximate aspects of this delegation picture, including federated learning \cite{nori2025federated}, constructivist AGI programmes \cite{thorisson2012,goertzel2014}, and active inference \cite{Friston2010,friston2013}. An expanded discussion (including AERA, Hyperon/ActPC-Chem, and links to scale-free active inference) is provided in the ESM (Supplementary Note~S6). \emph{Stack Theory} might also inform the design of `soft robots' \cite{man2019}. Soft robots are a proposed class of robots made of `soft' materials that adapt. I have formalised adaptive materials as adaptation lower in the stack, which can be used as a formal foundation for soft robotics. Self-assembling nanoparticle networks \cite{borghi2024_brainlikehw} are a step in this direction. They can be used to form reconfigurable logic gates \cite{paroli2023_receptron}, facilitating adaptation at very low levels of abstraction. Taken to its logical conclusion, future AI may more closely resemble artificial life \cite{suzuki2006,ikegami2008,ikegami2007simulating}. Finally, \emph{Stack Theory} may be extended to economic and organisational systems. For example, free-market trade could also be seen as a classic argument for delegation. As Hayek pointed out \cite{hayek1945use}, a great deal of relevant information is spread among a population, unavailable to a central planner. This motivates delegation of control, to shorten feedback loops. However, the present formalism treats resources implicitly via task correctness, so connecting these arguments to economic dynamics would require extending Stack Theory to explicit time-series \cite{bennett2026a} and resource accumulation. Additional speculative extensions and applications (including organisational and economic systems, AI safety regulation, and temporal extensions of stacks) are collected in ESM Note~S9.

In summary, I have applied \emph{Stack Theory} to explore how delegation of control affects adaptability. \emph{The Law of the Stack} shows how adaptability at higher levels of abstraction is constrained by adaptability at lower levels, and delegation is structurally necessary to minimise free energy. A practical design implication is that, when adaptability is desired, constraints should be weakened as far as possible without breaking the system's intended function. Constraints that are not strictly necessary relative to the design goal should be removed, though what counts as ``strictly necessary'' is relative to design goals and value judgements (See ESM Supplementary Note~S4.). Humans often prefer `additive' solutions \cite{adams2021}, meaning if a problem can be solved by adding a rule or subtracting a rule, we tend to add a rule. However \emph{The Law of the Stack} suggests a preference for subtractive solutions might be beneficial in many circumstances. The Law of the Stack might also be viewed as a formal justification of the classical principle of subsidiarity, often traced back to Aristotle, namely that higher levels of organization should coordinate and enable lower levels rather than replace functions they can adequately perform \cite{aristotle_politics}. It points out a principle common to many systems, formalised in terms of \emph{Stack Theory} and structurally tied to the FEP.
Hence this paper lays a foundation for a broad range of future theoretical and empirical research, showing how systems of any kind might be made more robust and adaptable.

\section*{Acknowledgements}
I thank the anonymous peer reviewers for their detailed and helpful feedback. It made a substantial difference. I also thank Anna Ciaunica, Peter Watts, Sean Welsh, Ricard Solé, Karl Friston, Antonio Damasio, Kingson Man, Paolo Milani, Ben Goertzel, Kristinn R. Thórisson, Elija Perrier and Noel Hinton for their support and feedback on this paper. 

\bibliographystyle{RS}
\bibliography{main_fixed}

\end{document}

%\linenumbers

\title{Electronic Supplementary Materials for Are Biological Systems More Intelligent Than Artificial Intelligence?}

\author{Michael Timothy Bennett}

\address{The Australian National University}

\subject{Hybrid Agency, Artificial Intelligence, Cyberphysical Systems}

\keywords{Stack Theory, delegation, w-maxing, adaptability, hybrid agency}

\corres{Michael Timothy Bennett\\
\email{m@michaeltimothybennett.com}}

\begin{abstract}
This is a collection of supporting material for the paper, including the complete mathematical proofs. To facilitate numbered references to theorems, it is marked as section 9 of the paper.
\end{abstract}

\maketitle

\providecommand{\st}[1]{#1} 

\section{Introduction}

\begin{claim}[Delegation constrains adaptability]\label{clm:delegation}
\end{claim}

\begin{claim}[Delegation is necessary for free energy minimisation]\label{clm:fep}
\end{claim}

\begin{claim}[Insufficient delegation can induce cancer-analogue fragmentation]\label{clm:cancer}
\end{claim}

\begin{claim}[Hybrid creation as maximally weak boundary-condition design]\label{clm:hybrid}
\end{claim}

\section{Abstraction, Distribution and Delegation} \label{sectionUpperBounds}

\subsection{Abstraction}

\subsection{Distribution}

\subsection{Delegated Control}

\subsection{Reasons to Suspect Delegation Facilitates Adaptation}

\section{What is Adaptation?}

\section{Formal Framework and Definitions}

\section{Delegated Control and Collective Identity}

\subsection{The Stack}

\subsection{Delegation and variational free energy}

\subsection{Human Abstractions}

\section{Cancer-Like Failure Modes} \label{sectionCancer}

\subsection*{Formal characterisation of the cancer-analogue condition}

\section{Hybrid Creation as Weak Boundary-Condition Design}

\section{Conclusions and Future Research} \label{sectionConclusion}

\section*{Acknowledgements}

\newpage
%\printbibliography
 \setcounter{page}{2}
%\begin{comment}
\section{Electronic Supplementary Materials}
\subsection{Mathematical Results}

\subsection*{The Law of the Stack}

\begin{theorem}[The Law of the Stack]
Assume the formal setting defined in the main text.
Finite vocabulary $v\subseteq P$, truth sets $T(\ell)=\bigcap_{p\in\ell} p$, language $L_v=\{\,\ell\subseteq v:T(\ell)\neq\emptyset\,\}$,
extensions $\Ext{\pi}=\{\,y\in L_v:\pi\subseteq y\,\}$, $v$-tasks $\gamma=\langle I_\gamma,O_\gamma\rangle$,
correct policies $\Pi_\gamma=\{\pi\in L_v: \Ext{I_\gamma}\cap \Ext{\pi}=O_\gamma\}$, and utility
$\epsilon(\gamma)=\max_{\pi\in\Pi_\gamma}(|\Ext{\pi}|-|O_\gamma|)$.
Let $\langle \lambda_0,\dots,\lambda_n\rangle$ be a stack (MLA) with abstractor $\mathfrak{f}$ and state
$\langle (\pi_0,\dots,\pi_n),(v_0,\dots,v_n)\rangle$ such that
$v_{i+1}=\mathfrak{f}(v_i,\pi_i)$ and $\pi_i\in\Pi_{\lambda_i(v_i)}$ for all $i<n$.

Fix any $i<n$, and define $\gamma_{i+1}=\lambda_{i+1}(v_{i+1})$ where $v_{i+1}=\mathfrak{f}(v_i,\pi_i)$.
Assume $\Pi_{\gamma_{i+1}}\neq\emptyset$ so that $\epsilon(\gamma_{i+1})$ is well-defined.
Then
\[
\epsilon(\gamma_{i+1}) + |O_{\gamma_{i+1}}| \;\le\; 2^{|\Ext{\pi_i}|}.
\]
Equivalently,
\[
|\Ext{\pi_i}|\;\ge\;\log_2\!\big(\epsilon(\gamma_{i+1})+|O_{\gamma_{i+1}}|\big).
\]
Thus, higher utility at layer $i\!+\!1$ requires (at least) a weaker policy at layer $i$
in the sense of larger extension cardinality.
\end{theorem}
\Intuition This theorem is the formal version of the delegation claim. It says that a layer can only support many distinct correct outputs if the realised policy one layer down leaves many compatible completions. The proof is a counting argument that converts this slack into an upper bound on $\epsilon(\gamma_{i+1})$.

\begin{proof}
We proceed by a sequence of elementary bounds that make the layer-to-layer dependency explicit.

\paragraph{Lemma 1 (Languages are bounded by the power set).}
For any finite vocabulary $v$, the language $L_v$ satisfies $|L_v|\le 2^{|v|}$.\\

\noindent \emph{Proof.} By definition, $L_v\subseteq \mathcal{P}(v)$ because every $\ell\in L_v$ is a subset
of $v$. Since $|\mathcal{P}(v)|=2^{|v|}$ for finite $v$, we have $|L_v|\le 2^{|v|}$. \hfill$\square$\\
\ProofIntuition A policy is a statement in the language, so counting statements bounds how many policies exist.

\paragraph{Lemma 2 (Extensions live inside the language).}
If $\pi\in L_v$, then $\Ext{\pi}\subseteq L_v$, hence $|\Ext{\pi}|\le |L_v|$.\\

\noindent\emph{Proof.} By definition $\Ext{\pi}=\{y\in L_v:\pi\subseteq y\}$, so $\Ext{\pi}\subseteq L_v$.
Taking cardinalities yields $|\Ext{\pi}|\le |L_v|$. \hfill$\square$\\
\ProofIntuition An extension only contains statements already in the language, so it cannot be larger than the language itself.

\paragraph{Lemma 3 (Utility implies a lower bound on language size).}
Let $\gamma$ be a $v$-task with $\Pi_\gamma\neq\emptyset$. Then
\[
\epsilon(\gamma)+|O_\gamma|\le |L_v|.
\]

\noindent\emph{Proof.} Because $v$ is finite, $L_v$ is finite, hence $\Pi_\gamma\subseteq L_v$ is finite. Since $\Pi_\gamma\neq\emptyset$, the maximum defining $\epsilon(\gamma)$ is attained by some $\pi^\star\in\Pi_\gamma$.
By definition,
\[
\epsilon(\gamma)=|\Ext{\pi^\star}|-|O_\gamma|\quad\Rightarrow\quad
\epsilon(\gamma)+|O_\gamma|=|\Ext{\pi^\star}|.
\]
By Lemma 2, $|\Ext{\pi^\star}|\le |L_v|$. Combining yields the claim. \hfill$\square$\\

\ProofIntuition Utility measures how many additional completions a best policy has beyond the correct outputs, so the total is bounded by the number of possible statements.

\paragraph{Lemma 4 (The abstractor output vocabulary is an image of an extension).}
Let $T:L_{v_i}\to P$ be the map $T(o)=\bigcap_{p\in o} p$. Under the stack (MLA) definition,
\[
v_{i+1}=\mathfrak{f}(v_i,\pi_i)=\{\,T(o): o\in \Ext{\pi_i}\,\},
\]
so $v_{i+1}$ is the image of $\Ext{\pi_i}$ under $T$. Consequently,
\[
|v_{i+1}|\le |\Ext{\pi_i}|.
\]

\noindent\emph{Proof.} This is just re-writing the definition of $\mathfrak{f}$ in functional form, and then applying
the general fact that for finite sets $A$ and any function $g$, $|g(A)|\le |A|$. \hfill$\square$\\
\ProofIntuition The abstractor creates one next-layer predicate per completion, so the induced vocabulary size is bounded by the number of completions.

\paragraph{Conclusion.}
Apply Lemma 3 to $\gamma_{i+1}$ in vocabulary $v_{i+1}$ to obtain
\[
\epsilon(\gamma_{i+1})+|O_{\gamma_{i+1}}|\le |L_{v_{i+1}}|.
\]
Apply Lemma 1 to bound $|L_{v_{i+1}}|\le 2^{|v_{i+1}|}$, giving
\[
\epsilon(\gamma_{i+1})+|O_{\gamma_{i+1}}|\le 2^{|v_{i+1}|}.
\]
Finally apply Lemma 4. It gives $|v_{i+1}|\le |\Ext{\pi_i}|$. Since $2^x$ is monotone increasing,
\[
2^{|v_{i+1}|}\le 2^{|\Ext{\pi_i}|}.
\]
Combining the last two displays yields
\[
\epsilon(\gamma_{i+1})+|O_{\gamma_{i+1}}|\le 2^{|\Ext{\pi_i}|},
\]
and the logarithmic form follows by applying $\log_2$ (also monotone). \hfill 
\end{proof}
\ProofIntuition The proof decomposes the bound into four combinatorial steps. Lemma 1 relates the number of policies to the number of statements. Lemma 2 turns a bound on the number of correct policies into a bound on achievable outputs and utility. Lemma 3 counts statements by the size of the induced vocabulary. Lemma 4 shows that the induced vocabulary is no larger than the number of completions of the realised lower-layer policy.\\

\noindent \textbf{Remark.} In the thesis version of this result, the intuition is phrased as follows. Larger vocabularies at layer
$i{+}1$ permit weaker (more general) policies, hence more adaptation at higher layers. The proof
here makes the dependency explicit by bounding the achievable utility at layer $i{+}1$ as a monotone
function of the extension size $|\Ext{\pi_i}|$ at layer $i$, without requiring any additional
monotonicity assumptions about $\epsilon(\lambda(\cdot))$ under set inclusion.

\subsection*{Grounding and compositional grounding}

The proof of Theorem 5.1 reasons about how a realised policy at one layer induces the next-layer vocabulary via the abstractor $\mathfrak{f}$.
It is also useful to make explicit how statements at a higher layer can be translated back into equivalent lower-layer statements.
This translation is not unique because each induced next-layer program can have multiple compatible completions.

\paragraph{Definition (A grounding map for one abstraction step).}
Fix a finite vocabulary $v\subseteq P$ and a policy $\pi\in L_v$.
Let $v'=\mathfrak{f}(v,\pi)=\{\,T(o)\mid o\in\Ext{\pi}\,\}$.
For each $p\in v'$ choose one completion $o_p\in\Ext{\pi}$ such that $T(o_p)=p$.
Define the grounding map $\mathrm{ground}_{\pi}\colon L_{v'}\to L_v$ by
\[
\mathrm{ground}_{\pi}(\ell)=\bigcup_{p\in \ell} o_p.
\]

\Intuition A higher-layer program $p\in v'$ represents the truth set of some completion compatible with $\pi$.
Grounding replaces each such higher-layer program by one concrete lower-layer completion that realises it.
The union then aggregates these lower-layer constraints into one lower-layer statement.

\paragraph{Lemma (Grounding preserves truth conditions for one step).}
For every $\ell\in L_{v'}$ we have $T(\mathrm{ground}_{\pi}(\ell))=T(\ell)$.

\begin{proof}
By definition,
\[
T(\mathrm{ground}_{\pi}(\ell))
=
\bigcap_{q\in \bigcup_{p\in \ell} o_p} q
=
\bigcap_{p\in \ell}\Big(\bigcap_{q\in o_p} q\Big)
=
\bigcap_{p\in \ell} T(o_p).
\]
By construction $T(o_p)=p$ for each $p\in v'$, so
\[
T(\mathrm{ground}_{\pi}(\ell))
=
\bigcap_{p\in \ell} p
=
T(\ell).
\]
\end{proof}

\ProofIntuition The truth set of a union of constraints is the intersection of the truth sets.
Each selected completion $o_p$ has truth set exactly $p$.
Intersecting over $p\in\ell$ therefore reproduces $T(\ell)$.

\paragraph{Definition (Compositional grounding across a stack).}
Fix a sequence of vocabularies and policies
\[
v^0,\dots,v^m
\qquad\text{and}\qquad
\pi^0,\dots,\pi^{m-1}
\]
such that $v^{i+1}=\mathfrak{f}(v^i,\pi^i)$ for each $i<m$.
Choose grounding maps $\mathrm{ground}_{\pi^i}\colon L_{v^{i+1}}\to L_{v^i}$ as above.
For any $\ell^m\in L_{v^m}$ define $\mathrm{Ground}_{0\leftarrow m}(\ell^m)\in L_{v^0}$ recursively by setting $g^m=\ell^m$ and
\[
g^i=\mathrm{ground}_{\pi^i}(g^{i+1})
\qquad\text{for } i=m-1,m-2,\dots,0.
\]
Then $\mathrm{Ground}_{0\leftarrow m}(\ell^m)=g^0$.\\

\Intuition This is iterated grounding.
You repeatedly replace a higher-layer statement by one lower-layer statement that has the same truth conditions.
After $m$ steps you obtain a statement in the base vocabulary that describes exactly the same set of states.

\paragraph{Theorem (Compositional grounding preserves truth conditions).}
For every $\ell^m\in L_{v^m}$ we have
\[
T(\mathrm{Ground}_{0\leftarrow m}(\ell^m))=T(\ell^m).
\]

\begin{proof}
By the one-step lemma, $T(g^i)=T(g^{i+1})$ for each $i<m$.
By induction, $T(g^0)=T(g^m)$.
Since $g^m=\ell^m$ and $\mathrm{Ground}_{0\leftarrow m}(\ell^m)=g^0$, the claim follows.
\end{proof}

\ProofIntuition Each grounding step preserves truth conditions.
Composing truth-preserving maps preserves truth conditions.\\

\noindent\textbf{Remark.}
Grounding depends on the choice of completions $o_p$ at each step.
Different choices can produce different lower-layer statements with the same truth conditions.

\subsection*{Viability, lived history, and variational free energy}

\paragraph{Setup.}
Let $\mu$ be a task representing the set $O_\mu$ of all viable behavioural continuations for an organism, as shaped by natural selection.
Let $\mathfrak{h}$ be the task representing the organism's lived history up to time $t$.
If the organism remains alive at time $t$, then $\mathfrak{h}$ is consistent with viability, and it is natural to treat $\mathfrak{h}$ as a child task of $\mu$.
In the present language (and using the monotonicity of policy sets under the task order), this means that every policy that is correct for viability is also consistent with the lived history up to time $t$, so $\Pi_{\mu} \subseteq \Pi_{\mathfrak{h}}$.
Fix a planning horizon and write $\Omega$ for the set of possible future continuations at that horizon, expressed at the relevant abstraction layer.

Given a policy $\pi$, write
\[
\Omega_\pi = \Ext{\pi} \cap O_\mu.
\]
This is the set of viable continuations that remain compatible with $\pi$.
We will only consider policies with $\Omega_\pi\neq\emptyset$, so the logarithms below are well-defined.
Define the viability weakness score
\[
w(\pi\mid \mathfrak{h},\mu) = \log_2 |\Omega_\pi|.
\]
A viability w-maxing agent selects $\pi^\star\in \Pi_{\mathfrak{h}}$ with $\Omega_{\pi^\star}\neq\emptyset$ maximising $w(\pi\mid \mathfrak{h},\mu)$.
Intuitively, among policies that explain the past, it prefers the one that leaves the largest set of viable futures.

Now connect this to the variational free energy functional.
Let $q(\omega)$ be any distribution over future continuations $\omega\in \Omega$.
Let the viability preference be the uniform distribution over viable continuations
\[
p_\mu(\omega) = \frac{1}{|O_\mu|}\,\mathbf{1}\{\omega\in O_\mu\}.
\]
Define the (base-2) free energy of $q$ relative to $p_\mu$ by
\[
F_2(q) = \mathbb{E}_{q}[-\log_2 p_\mu(\omega)] - H_2(q),
\]
where $H_2(q)$ is Shannon entropy (in bits).

For each policy $\pi$ with $\Omega_\pi\neq\emptyset$, consider the maximum entropy distribution supported on $\Omega_\pi$
\[
q_\pi(\omega) = \frac{1}{|\Omega_\pi|}\,\mathbf{1}\{\omega\in \Omega_\pi\}.
\]

\noindent\textit{Formally, the following theorem shows that under a uniform viability prior, free energy minimisation is exactly equivalent to maximising the log-count of viable future continuations (viability weakness).}

\paragraph{Theorem (Equivalence of viability w-maxing and free energy minimisation).}
Assume the viability prior is uniform on $O_\mu$ as above.
Then for any $\pi\in \Pi_{\mathfrak{h}}$ with $\Omega_\pi\neq\emptyset$,
\[
F_2(q_\pi) = \log_2 |O_\mu| - \log_2 |\Omega_\pi|.
\]
Consequently, over $\pi\in \Pi_{\mathfrak{h}}$ with $\Omega_\pi\neq\emptyset$, maximising $w(\pi\mid \mathfrak{h},\mu)$ is equivalent to minimising $F_2(q_\pi)$.

\begin{proof}
For $\omega\in O_\mu$ we have $-\log_2 p_\mu(\omega)=\log_2 |O_\mu|$.
Since $q_\pi$ is supported on $\Omega_\pi\subseteq O_\mu$, it follows that
\[
\mathbb{E}_{q_\pi}[-\log_2 p_\mu(\omega)] = \log_2 |O_\mu|.
\]
Also, since $q_\pi$ is uniform on a set of size $|\Omega_\pi|$, its entropy is $H_2(q_\pi)=\log_2 |\Omega_\pi|$.
Therefore
\[
F_2(q_\pi)=\log_2 |O_\mu|-\log_2 |\Omega_\pi|,
\]
and the equivalence follows because $\log_2 |O_\mu|$ is constant with respect to $\pi$.
\end{proof}

\ProofIntuition Under a viability prior that treats all viable futures as equally acceptable, free energy minimisation reduces to maximum entropy. Maximum entropy under the support constraint imposed by a policy means preferring policies that leave more viable continuations.

\noindent\textit{For completeness, the next lemma records the standard variational bound that underwrites free energy as an upper bound on surprisal.}

\paragraph{Lemma (Variational free energy bounds surprisal).}
Let $p_\mu(\omega,s)$ be any generative model over future continuations $\omega$ and latent variables $s$, and let $q(s)$ be any density.
Define
\[
F_2(q,\omega) = \mathbb{E}_q[\log_2 q(s) - \log_2 p_\mu(\omega,s)].
\]
Then
\[
F_2(q,\omega)= \mathrm{KL}(q(s)\Vert p_\mu(s\mid \omega)) - \log_2 p_\mu(\omega) \ge -\log_2 p_\mu(\omega).
\]
In particular, minimising $F_2(q,\omega)$ over $q$ yields $-\log_2 p_\mu(\omega)$.

\begin{proof}
Write $p_\mu(\omega,s)=p_\mu(s\mid \omega)p_\mu(\omega)$ and substitute into the definition of $F_2(q,\omega)$.
The resulting expression is $\mathrm{KL}(q(s)\Vert p_\mu(s\mid \omega)) - \log_2 p_\mu(\omega)$.
Since KL divergence is non-negative, the bound follows.
\end{proof}

\noindent\textbf{Remark.}
The uniform viability prior is a stylised proxy for natural selection.
More structured preferences correspond to non-uniform $p_\mu$ and to the full Active Inference objective based on expected free energy.
Even in that more general case, the same decomposition holds, with a fit term and an entropy term.
The mapping above shows how a weakness style objective can be recovered as the entropy seeking component when viability is treated as a hard constraint. Combined with The Law of the Stack, we get a delegation bottleneck. The entropy of viable futures at a layer is upper-bounded by the slack (weakness) available one layer down (see Corollary 5.1 in the main text).

\subsection*{Hybrid case study formalisation (Pong organoid)}
Let $C$ be a set of cells.
For each cell $i\in C$, let $\Pi_i$ denote the set of viable policies for that cell under some interface and environmental conditions.
The set of viable collective policies is
\[
\Pi_C=\bigcap_{i\in C}\Pi_i.
\]
A conventional computer interface imposes boundary conditions $c$ that restrict which policies remain viable.
Write this restriction as an operator $R_c$ so that $\Pi_i^{(c)}=R_c(\Pi_i)$ and $\Pi_C^{(c)}=\bigcap_{i\in C}\Pi_i^{(c)}$.
The hybrid design objective is to choose $c$ so that $\Pi_C^{(c)}\neq\emptyset$.
The surviving collective policies should also implement the target interface behaviour.
In the Pong example, this means that the set of achievable outputs at the interface includes the correct output set $O_{\mathrm{pong}}$ for the Pong task.
In this view, building the hybrid is a boundary-condition design problem.
Impose the weakest boundary conditions that make $O_{\mathrm{pong}}$ achievable while preserving non-empty intersection, which is collective identity.\\

\Intuition The intersection $\Pi_C$ expresses whether the parts can act as a coherent whole.
Changing the interface changes which policies remain viable for each part and therefore changes the intersection.
The design goal is to shrink the intersection just enough to realise the desired task without destroying the intersection entirely.
\subsection{Supplementary Notes and Extended Discussion}

\subsection*{S1. Working definition of intelligence}
There are many definitions of intelligence, and the merits of each are beyond the scope of this paper. Mine is based on \cite{bennett2025thesis}, which frames intelligence as the ability to complete a wide range of tasks, which is sample and energy efficiency in adaptation. It is similar to Wang's definition of intelligence as adaptation with limited resources \cite{wang2019}, and Hutter's earlier definition of intelligence as `the ability to satisfy goals in a wide range of environments'. Mainstream AI often operationalizes `intelligence' as general-purpose task performance. By that standard, one may argue that current large language models are `more intelligent' than many biological systems. However that definition is fundamentally flawed, as it fails to express the comparative efficiency with which humans acquire general purpose competence \cite{bennett2024a,wang2019,chollet2019}. In theory even a lookup table or rote-learning response can possess general purpose competence, given enough resources. This paper instead treats intelligence as the cost of adaptation (sample and energy efficiency), and the results should be read accordingly.

\subsection*{S2. Levels of abstraction beyond computing}
The notion of levels of abstraction is more typically used in pure computer science settings, but can be applied to any system \cite{bennett2025thesis}. A higher level of abstraction is the behaviour of the layers below. For example, an organ like a heart is a behaviour of cells, human is a behaviour of some organs, and language is a behaviour of humans.

\subsection*{S3. Weakness and w-maxing (extended intuition)}
Weakness is formally defined in the appendix. Related work proved w-maxing to be optimal regardless of the abstraction layer \cite{bennett2025e,bennett2023b,bennett2024a,bennett2025thesis}. In narrative terms, assume two policies $x$ and $y$ both constrain the system to examples of ``correct'' behaviour given past inputs, but differ in weakness. If $x$ is weaker than $y$, that means it commits to less, but achieves the same end. For example, if I assume the lawyer was paid at least $\$5$, I am more weakly constrained than if I assume that lawyer was paid at least $\$6$.

This example also helps separate weakness from simplicity. The difference between $\$5$ and $\$6$ is not primarily about how hard the rule is to write down. It is about how many further commitments remain compatible with the rule, given the agent's available distinctions. Weakness is defined relative to an abstraction layer. The layer fixes what counts as a meaningful refinement because it fixes what the system can jointly discriminate and jointly realise. If the embodiment changes, then the induced vocabulary changes, and so does the completion structure that weakness counts.

Many familiar heuristics can be seen as special cases of the same idea once this completion structure is made explicit. If one treats all completions that remain possible at a layer as equally plausible, then choosing a correct policy that maximises $|\Ext{\pi}|$ is equivalent to maximising entropy over viable futures inside that layer. This parallels least commitment and version space intuitions, but it is stated directly in terms of the agent's embodied language rather than in terms of an encoding chosen by an observer. That is what lets weakness be used across layers in Stack Theory, where a policy at one layer induces the next layer vocabulary via the abstractor map.

In the language of agency, weakness can be read as a quantitative proxy for autonomy at a layer. A weaker policy leaves more degrees of freedom (more ``ways of being right'') while still satisfying the same observed constraints. In multiscale and hybrid systems this autonomy matters because it increases the chance that component policies remain compatible with a higher-level collective policy, preserving collective identity.

As an epistemological razor, this means ``explanations should be no more specific than necessary'' or alternatively ``rules should constrain no more than necessary'' (see Bennett's Razor in \cite{bennett2023b}). This is a bit like the principle of maximum entropy \cite{jaynes1957a,jaynes1957b}, difference being that w-maxing also accounts for the abstraction layer, which is important because abstraction can render complexity subjective \cite{bennett2024b}.

For a rigorous definition of learning and efficiency, see definitions in the online repository for the thesis \cite{bennett2024appendices}. It includes proofs showing w-maxing is both necessary and sufficient to maximise the probability of `generalising' from one task to another, and thus adaptability.

\subsection*{S4. Normativity and ``strict necessity'' in design}
As Hume argued, a description of what ought to be cannot be derived from only descriptions of what is. ``Strictly necessary'' is a value judgement, or an implicit choice of what ought to be. In the thesis \cite{bennett2025thesis} on which this is based this ought from which other descriptions of ought follow is derived from the state transitions of the environment. These transitions preserve some aspects of that environment, while destroying others. Here however I am more concerned with the design of systems which serve a human purpose. Strictly necessary is a matter of desired functionality.

\subsection*{S5. Conceptual motivation for the formal setting}
Stack Theory frames the description of an objective reality in terms of choices of subjective lens (an abstraction layer) \cite{bennett2025thesis}. The full justification for these definitions is out of scope and addressed at length elsewhere \cite{bennett2024a,bennett2025thesis,bennett2025e}. However I will provide a brief overview. To make claims that hold regardless of the abstraction layer, we must rely only on what holds for every abstraction layer. From our subjective perspective we can't know what is at the bottom of our stack (it may be infinite). I call this bottom layer the \textbf{environment}. Whatever the environment is, there must be change or difference for there to be anything at higher levels of abstraction. Hence there must be a set of \textbf{states} where each state is one point of difference (Nothing is assumed about what these states might ontologically be or epistemologically contain.). Because these states must be mutually exclusive, this frames time as difference. Each possible sequence of states is a timeline or possible world. Truth is defined ontologically, relative to states. A set of states is effectively a set of truth conditions, so a possible fact is called a declarative \textbf{program}, and a program is \texttt{true} given a state $\phi$ if it contains that state. Every aspect of the environment is just a set of programs. Everything that exists is a set of facts, and each timeline preserves different sets of facts. A timeline is like an opinion. As time progresses some aspects of the environment are preserved and others are destroyed. Each timeline preserves different aspects, expressing a different ``ought'' which related work uses to explain the origins of goal-directed behaviour \cite{bennett2025thesis}. If the set of states is infinite then there are infinite possible worlds. Assume I have a robotic system. I get an \textbf{abstraction layer} for the system by choosing a finite aspect $\mathfrak{v}$ of programs called a \textbf{vocabulary}. Everything the abstraction layer can express is a subset of its vocabulary. For example if a system has an arm it can raise or lower, then there are programs in $\mathfrak{v}$ which correspond to actuation of the arm in some form, though it need not be neat symbolic representations like 'raise arm' or 'lower arm'. Since states are mutually exclusive, $\mathfrak{v}$ gives us an embodied formal language. I do not mean like the words humans use to communicate. I mean a language of embodiment, where every statement in that language \textit{constrains} rather than \textit{describes} the system, like the aforementioned boundary conditions on slime mold \cite{nakagaki2000} (Using this frame cognition can be formalised in terms that align with both enactivist \cite{thompson2007} and pancomputationalist \cite{piccinini2021} perspectives. Hence, cognition within Stack Theory is also called Pancomputational Enactivism \cite{bennett2024a}.)

\subsection*{S6. Extended related work on delegated software architectures and Active Inference/FEP}
There are examples of systems that take something like the present approach, albeit within the confines of software abstraction layers.

Federated machine learning delegates control to some extent (e.g.\ shifting data ownership and parts of optimisation to edge devices), which can improve privacy and efficiency \cite{nori2025federated}. Constructivist theories of artificial general intelligence emphasise multilayered self-organisation at the level of software while acknowledging at least some of the constraints imposed by hardware \cite{thorisson2012,goertzel2014}. The Auto-catalytic Endogenous Reflective Architecture (AERA) is a goal-driven ampliative reasoning system \cite{nivel2013} based on constructivist principles that relies on causal knowledge representation to achieve autonomy \cite{thorisson2018cumlearncause}, ``growing'' knowledge in a collective meta-model through cumulative continuous learning \cite{thorisson2019cumlearning}, and explicitly targeting adaptability \cite{thorisson2014autonomousNatCom,thorisson2020,eberding2024,sheikhlar2024} (See also \url{http://www.openaera.org}.). ActPC-Chem \cite{goertzel2024} is a recent theory behind the distributed Hyperon architecture \cite{goertzel2023} that was influenced by theoretical and methodological insights from AERA.

\paragraph{Active Inference/FEP in relation to MCL.}
Active Inference (and the Free Energy Principle, FEP) provides a normative account of perception and action in which an agent maintains a generative model and selects actions to minimise variational free energy (and, by extension, expected free energy) \cite{Friston2010,friston2013,parr2024}. The present paper addresses a different axis. It states a structural constraint on \emph{stacks of abstraction layers}, namely that what can be achieved at layer $i{+}1$ (utility $\epsilon(\gamma_{i+1})$) is bounded by the extension (weakness) of the realised policy at layer $i$ (Theorem 5.1). In this sense, MCL is about \emph{where} adaptation is implemented (and how much latitude lower layers retain), rather than about the particular inferential objective used within a layer. An Active Inference agent implemented on a predominantly fixed computational substrate can still be bottlenecked by static lower layers. Conversely, a delegated multiscale substrate can in principle host a range of learning rules, including Active Inference.

\paragraph{Potential complementarity and open questions.}
Several lines of work suggest points of contact, without implying formal equivalence. ActPC-Chem frames delegation in explicitly Active Inference terms \cite{goertzel2024,Friston2010,friston2013}. ``Scale-free'' active inference proposes that similar dynamical principles repeat across scales \cite{friston2023b,friston2024a}, which is conceptually close to the MCA picture and to the idea that policy alignment must be maintained across layers to preserve collective identity. Delegated \cite{ororbia2022,ororbia2023,ororbia2023b} or federated \cite{friston2024b} Active Inference approaches may therefore be promising candidates for building efficient hybrid systems at scale. Separately, work on ``mortal computation'' seeks to approach thermodynamic limits (e.g.\ the Landauer bound) by reducing memory and processor separation and associated bottlenecks \cite{ororbia2024mortalcomputationfoundationbiomimetic,aurell2012,zou2021}, which resonates with the present paper's emphasis on efficiency and with the intuition that pushing adaptation downward can reduce overhead. However, the theories still differ in their primitives (e.g.\ tasks and weakness versus generative models and free energy). Even so, a simple mapping is available when viability is treated as a hard constraint and preferences are uniform across viable futures, see the Mathematical Results subsection ``Viability, lived history, and variational free energy''. A deeper identification in the general non-uniform case remains open.

\subsection*{S7. Mission command example (Nordbat 2)}
A well documented example of the effectiveness of mission command in facilitating adaptation is Nordbat 2, a joint peacekeeping force deployed to Bosnia in the early 90s. Mission command allowed individual units to deviate from convention, adapt to local conditions and adopt a more `trigger-happy' disposition that proved to be substantially more effective than the more rigid, rule based approach taken by the Dutch forces operating in the same region \cite{ingesson2016,ingesson2017}.

Strategic-level intent and a small set of constraints are communicated downward, while tactical-level decision-makers fill in the remaining degrees of freedom on the ground. In this reading, the bridge between leadership and soldiers is not a detailed plan but a constraint system that keeps local improvisation aligned with higher-level purpose. The purpose of this note is not to treat Nordbat~2 as empirical evidence for Theorem 5.1, but to provide an intuition-building picture of what Stack Theory means by delegation \cite{ingesson2016,ingesson2017}.

Mapped to the formalism, the strategic layer specifies a parent task $\omega$ (intent, mandate, and coordination requirements), while the tactical layer faces a family of child tasks $\alpha\sqsubset\omega$ that vary with local information and contingencies. Mission command enlarges the set of admissible completions by avoiding over-specification of the realised policy. It preserves a larger extension $|\Ext{\pi_i}|$ at the lower layer, enabling more distinct higher-layer behaviours to be expressed one layer up (Theorem 5.1).

For hybrid and cyberphysical design, the analogy is that effective ``control'' can often be achieved by engineering boundary conditions (interfaces, reward signals, resource limits, safety constraints) rather than by prescribing micro-actions. Over-prescriptive constraints risk collapsing the intersection of viable component policies (loss of collective identity), whereas intent-plus-boundaries can preserve slack for local adaptation while maintaining coherence.

\subsection*{S8. Candidate empirical and simulation tests}
The main text uses analogies (mission command, economic delegation, and distributed swarms) as intuition-building thought experiments rather than as evidence. It is primarily conceptual.
This note lists candidate tests that target the core claim of Theorem 5.1.
The goal is to compare systems under controlled variation of where adaptation is permitted in the stack.

\paragraph{1. Agent-based simulations with tunable delegation.}
Construct an agent-based model of a collective.
Include a delegation parameter that controls whether decisions are taken locally or routed through higher-level coordination.
Expose the collective to abrupt environmental shifts.
Measure sample efficiency, energy use, robustness, and identity preservation.
Test whether deeper delegation improves efficiency in the sense predicted by reduced weakness of lower-layer realised policies.

\paragraph{2. Comparative benchmarking across stack depth.}
Compare learning systems that differ only in where adaptation is permitted.
Hold the top-layer objective fixed and vary whether lower layers are static or reconfigurable.
For example, compare fixed-substrate deployment against deployment that allows adaptation of firmware, routing, or morphology.
Measure adaptation speed and compute the effective induced vocabulary size at each layer when possible.
This isolates the stack-level delegation hypothesis from algorithm choice at the top.

\paragraph{3. Hybrid boundary-condition sweeps.}
For hybrid agents such as simulated mixtures of human and AI agents or engineered organoids, treat alignment as hybrid agent design.
Sweep boundary conditions that prune low-level policy spaces.
Quantify the trade-off between task performance at the interface and preservation of non-empty collective policy intersection.
Look for phase transitions where the intersection collapses, which corresponds to loss of collective identity.

\paragraph{4. Interaction with Active Inference/FEP.}
Implement Active Inference style control at one or more layers.
Run the same comparison in an architecture that differs only in whether lower layers can adapt.
This disentangles gains due to the inferential objective from gains due to deeper delegation.

\subsection*{S9. Extended applications and speculative directions}
This note collects material that is peripheral to the main theorem but may be useful for future-work framing. These applications are intended as design hypotheses rather than as evidence for
Theorem 5.1.

\paragraph{Organisational and economic systems.}
Free-market trade is a great example of how delegation can facilitate adaptability in a large collective. As Hayek pointed out \cite{hayek1945use}, not all of the information is available to a central planner and so control is best delegated to the individual. However, the accumulation of resources within my formalism is implicit, rather than explicit. I have been concerned only with whether a system satisfies whatever arbitrary notion of correct I choose (i.e. does it achieve homeostatic and reproductive goals in whatever amount of time I care about?), rather than how it \textit{evolves} over time as a dynamical system. In other words, I have concerned myself only with whether a system arrives on time at the destination I want, not the journey it took to get there. Future research might extend \emph{Stack Theory} to formalise time-series, or the accumulation of various resources. For example, a stack could represent a sequence of tasks over time where higher layers are later in time. Earlier I mentioned that task correctness can denote more holistic notions of emergent features robust to lower-level variation. A time-series model could describe the conditions under which such invariants emerge. It could be used to explore how the delegation of control, distribution of work and the accrual of resources change over time. For example under some conditions a free market may degrade into a maladaptive, top-down system. Those parts deprived of control or resources may then not be able to construct a policy that aligns with the larger whole, and so splinter off. To preserve the integrity of the system, policies which might support the desired invariants could be modelled and evaluated. Policy could be located along axes of distribution and delegation, like an alternative political compass. In a simple zero-sum game like a race, one example of such a policy is adaptive rubber banding \cite{mi2022}. A game mechanic (e.g. in Mario Kart) where success accrues risk as well as reward, keeping the game competitive.

\paragraph{Regulation and over-constraint risks.}
The cancer analogy suggests a general caution. Guardrails that over-constrain a system may
precipitate the failure mode they aim to prevent by collapsing the intersection of viable policies. For example, a survey of 3,306 Germans found that restrictions promoting carbon-neutral lifestyles would trigger strong negative responses because they `restrict freedom' \cite{schmelz2025}. 
This may be most relevant to proposals for safeguarded AGI and related governance approaches
\cite{dalrymple2024b,dalrymple2024a,anthropic2023,goertzel2014}, given the formal results pertain specifically to a generalisation-optimal model of intelligence \cite{bennett2025e}. These remarks are speculative and are intended as design hypotheses rather than claims about current regulatory regimes. From a philosophical perspective, the Law of the Stack provides a formal realization of the principle of subsidiarity in stack theoretic terms. Let $\{\lambda_i\}_{i=0}^n$ denote a stack of abstraction layers with delegation from $\lambda_{i+1}$ to $\lambda_i$. The Law of the Stack implies that higher layers preserve system viability by delegating control to lower layers whenever possible, intervening only to maintain global coherence. In this sense, optimal organization corresponds to maximizing weakness at higher layers, that is, minimizing unnecessary constraints on the behavior space of lower layers while retaining sufficient structure to ensure stack integrity. This formalizes the classical insight that higher levels of organization exist to enable rather than replace the agency of lower levels \cite{aristotle_politics,aquinas_de_regno_leonine}.

\bibliographystyle{RS}
\bibliography{main_fixed}

%\end{comment}